\documentclass{article}

\usepackage{arxiv}

\usepackage[utf8]{inputenc} 
\usepackage[T1]{fontenc}    
\usepackage{hyperref}       
\usepackage{url}            
\usepackage{booktabs}       
\usepackage{amsfonts}       
\usepackage{nicefrac}       
\usepackage{microtype}      
\usepackage{lipsum}

\usepackage{amsmath}
\usepackage{graphicx}

\title{Interpretable Vision Transformers in Image Classification via SVDA}

\author{
  Vasileios Arampatzakis\\
  Dept. Electrical and Computer Engineering\\
  Democritus University of Thrace\\
  Athena Research Center\\
  University Campus at Kimmeria\\
  67100 Xanthi, Greece\\
  \texttt{vaarampa@ee.duth.gr} \\
   \And
 George Pavlidis\\
Athena Research Center\\
University Campus at Kimmeria\\
67100 Xanthi, Greece\\
  \texttt{gpavlid@athenarc.gr} \\
  \And
Nikolaos Mitianoudis\\
Dept. Electrical and Computer Engineering\\
Democritus University of Thrace\\
University Campus at Kimmeria\\
67100 Xanthi, Greece\\
  \texttt{nmitiano@ee.duth.gr} \\
    \And
Nikos Papamarkos\\
Dept. Electrical and Computer Engineering\\
Democritus University of Thrace\\
University Campus at Kimmeria\\
67100 Xanthi, Greece\\
  \texttt{papamark@ee.duth.gr} \\
}

\begin{document}
\maketitle

\begin{abstract}
Vision Transformers (ViTs) have achieved state-of-the-art performance in image classification, yet their attention mechanisms often remain opaque and exhibit dense, non-structured behaviors. In this work, we adapt our previously proposed SVD-Inspired Attention (SVDA) mechanism to the ViT architecture, introducing a geometrically grounded formulation that enhances interpretability, sparsity, and spectral structure. We apply the use of interpretability indicators---originally proposed with SVDA---to monitor attention dynamics during training and assess structural properties of the learned representations. Experimental evaluations on four widely used benchmarks---CIFAR-10, FashionMNIST, CIFAR-100, and ImageNet-100---demonstrate that SVDA consistently yields more interpretable attention patterns without sacrificing classification accuracy. While the current framework offers descriptive insights rather than prescriptive guidance, our results establish SVDA as a comprehensive and informative tool for analyzing and developing structured attention models in computer vision. This work lays the foundation for future advances in explainable AI, spectral diagnostics, and attention-based model compression.
\end{abstract}


\section{Introduction}
Vision Transformers (ViTs) have demonstrated remarkable success in image classification tasks, often surpassing conventional convolutional architectures in various benchmarks \cite{dosovitskiy2020image}. However, despite their effectiveness, ViTs and other Transformer-based models face a persistent challenge: the lack of interpretability in their attention mechanisms. The standard self-attention formulation, based on dense dot-product interactions, often produces attention maps that are opaque and difficult to interpret. These patterns frequently lack spatial or semantic structure, limiting their utility in explaining model behavior. Prior work has shown that such maps may not reliably reflect the reasoning behind predictions \cite{wiegreffe2019attention}, and visualization tools alone often fail to provide actionable insight. The absence of semantically structured attention maps hinders both model interpretability and robustness across a wide range of tasks, from standard image classification benchmarks to high-stakes decision-making domains \cite{rudin2019stop}. Addressing this gap requires attention mechanisms that go beyond surface-level visualizations to offer geometric regularity and semantically meaningful structure.

\textbf{Motivation.} The interpretability of attention patterns has emerged as a key concern in both academic and applied settings. Studies have shown that attention maps may not reliably align with the model's decision rationale \cite{wiegreffe2019attention}, and popular visualization tools often fail to capture meaningful structure. In image classification, this undermines the potential for attention-based models to support tasks that require transparency, human oversight, or semantic reasoning. More broadly, in high-stakes domains such as healthcare, autonomous systems, and scientific discovery, reliance on black-box attention mechanisms poses ethical and practical risks \cite{rudin2019stop}. \textit{These concerns highlight the need for alternative formulations of attention that promote both transparency and structured behavior.}

\textbf{Our Approach.} To address this challenge, we propose the integration of our previously introduced SVD-inspired Attention (SVDA) mechanism \cite{arampatzakis2025geometry} into the Vision Transformer architecture. SVDA replaces the standard dot-product attention with a geometrically structured formulation that decouples directional information from spectral significance. It applies soft-orthonormal projections to the query and key spaces, combined with a learned diagonal matrix $\Sigma$ that modulates the attention interaction based on the spectral importance of each latent dimension. \textit{This structure induces attention maps that are inherently more focused, sparser, and interpretable, while preserving the original model architecture, the computational complexity, and the attained accuracy}. In addition to improving attention structure, SVDA enables a set of descriptive indicators that quantify the internal dynamics of attention layers. These indicators offer a principled diagnostic framework for analyzing attention interpretability, alignment, and information concentration across layers and heads.

\textbf{Scope and Contribution.} This paper explores the application of SVDA to image classification across four widely used datasets: CIFAR-10, CIFAR-100, FashionMNIST, and ImageNet-100. Through extensive experiments, we demonstrate that SVDA-based ViTs preserve competitive classification accuracy, while producing attention patterns that are more structured and interpretable. Additionally, we quantify attention behavior using spectral diagnostics and show improved robustness to perturbations. Our key contributions are:
\begin{itemize}
    \item The adaptation of SVDA to Vision Transformers, introducing geometric and spectral constraints for improved interpretability.
    \item A comparative evaluation on four standard benchmarks, showing that SVDA preserves performance while enhancing attention structure.
    \item The deployment of interpretability indicators as descriptive tools to assess internal attention dynamics and sparsity.
\end{itemize}

The results highlight the potential of SVDA as an interpretable and geometrically structured attention mechanism for image classification, bridging the gap between spectral theory and practical Transformer architectures. Our work introduces a principled framework for understanding and evaluating structured attention models, supported by spectral indicators that quantify interpretability and sparsity. These contributions lay the foundation for prescriptive attention design, where attention structures are explicitly optimized for transparency, and motivate further research in explainable AI, attention regularization, and spectral model compression.

\section{Related Work}

Several recent efforts have focused on enhancing interpretability in Vision Transformers \textit{by modifying how attention is visualized or trained}. For instance, Chefer~\emph{et al.}~\cite{chefer2021transformer} proposed gradient-based relevance propagation to improve the faithfulness of attention explanations in ViTs. Similarly, ProtoViT~\cite{ma2024protovit} integrates prototype-based reasoning to support case-based explanations, enabling ``this looks like that'' interpretability. IA-ViT~\cite{qiao2023interpretabilityaware} modifies the training objective to encourage inherently interpretable attention heads. These approaches emphasize interpretability either post hoc or through auxiliary training objectives, but they do not structurally alter the attention mechanism itself.


In parallel, several approaches have explored the \textit{injection of spectral and geometric structure} into attention mechanisms. SpecFormer~\cite{wang2023specformer} imposes spectral anisotropy regularization on attention maps through eigen-decomposition, promoting structured attention patterns across heads. Singularformer~\cite{guo2023singularformer} learns a neural approximation of the SVD of the attention matrix, enabling linear-time computation while preserving global dependencies. Cosformer~\cite{qin2022cosformer} replaces the dot-product kernel with cosine similarity to encourage directional alignment and improve efficiency. Primal-Attention~\cite{he2023primalattention} decomposes self-attention in the primal token space using a kernelized SVD formulation, enhancing representational capacity with theoretical guarantees. While each method introduces valuable structural priors, none provide a unified decomposition that explicitly separates directional and spectral contributions, nor do they furnish interpretable structural diagnostics for attention analysis.

Beyond structural modification, some works have aimed to \textit{develop metrics that quantify attention behavior}. Abnar and Zuidema~\cite{abnar2020quantifying} proposed attention flow as a means to measure information propagation across layers, but their approach is not designed to capture structural clarity, sparsity, or interpretability in a principled way. Our work introduces six structural indicators---spectral entropy, effective rank, angular alignment, selectivity index, spectral sparsity, and perturbation robustness---that together provide a descriptive framework for analyzing attention dynamics. These metrics quantify focus, diversity, alignment, and stability at the head and layer level, enabling principled diagnostics of interpretability and structural behavior in Vision Transformers. 

This paper builds on our prior work introducing SVDA~\cite{arampatzakis2025geometry}, which provided a theoretical formulation of structured attention based on soft-orthonormal projections and spectral modulation. In the current study, we adapt this formulation to the Vision Transformer architecture and apply it to image classification tasks. Unlike earlier works that treat attention as a black-box or seek to interpret it only after training, SVDA embeds interpretability into the attention computation itself. Moreover, it enables the use of principled structural diagnostics during both training and inference, paving the way for prescriptive design of transparent, structured attention models in computer vision.

\section{SVDA in Vision Transformers}

This work investigates the empirical impact of integrating SVDA~\cite{arampatzakis2025geometry} into Vision Transformers (ViTs) for image classification. SVDA introduces a principled decomposition of attention by disentangling geometric directionality from spectral importance. Inspired by the structure of the Singular Value Decomposition (SVD), SVDA replaces the standard dot-product attention with a formulation that incorporates $\ell_2$-normalized query and key projections, along with a learned diagonal spectral matrix $\Sigma$ per attention head. The core attention computation in SVDA is defined as:
\begin{equation}
\label{eq:attention}    
A \sim Q \Sigma K^\top
\end{equation}
where $Q, K \in \mathbb{R}^{n \times d_k}$ are the query and key matrices with row-wise $\ell_2$ normalization, and $\Sigma \in \mathbb{R}^{d_k \times d_k}$ is a learned diagonal matrix that modulates the relative importance of latent dimensions. This expression reflects the structural form of SVDA attention up to softmax scaling and head-wise normalization, and highlights the decomposition into directional and spectral components that enables fine-grained, interpretable attention behavior.

To ensure a direct comparison with the standard ViT architecture~\cite{dosovitskiy2020image}, our implementation follows the original embedding pipeline: each input image is divided into non-overlapping patches of size $P \times P$, which are flattened into vectors and passed through a learned linear projection to produce patch embeddings. A learnable [CLS] token is prepended, and fixed-length learnable positional embeddings are added to retain spatial order. The resulting token sequence is then processed by a Transformer encoder. The only deviation from the original ViT baseline is the replacement of the self-attention mechanism with SVDA, as defined in Eq.~\ref{eq:attention} and detailed in~\cite{arampatzakis2025geometry}.

\section{Experimental Setup}

\subsection{Datasets and Preprocessing}

We conducted experiments on four image classification benchmarks of increasing complexity: FashionMNIST, CIFAR-10, CIFAR-100, and ImageNet-100. Each dataset is processed using standard training/test splits and minimal preprocessing consistent with prior literature:
\begin{itemize}
    \item FashionMNIST: 28$\times$28 grayscale images of clothing items across 10 classes. Augmentations include random cropping (padding=4), horizontal flipping, and random rotation within $\pm$15°.
    \item CIFAR-10 / CIFAR-100: 32$\times$32 RGB images from 10/100 classes. Augmentations include random cropping (padding=4), horizontal flipping, and random rotation within $\pm$15°.
    \item ImageNet-100: A curated 100-class subset of ImageNet-1K, with images resized to 224$\times$224 and augmented with horizontal flipping. 
    \item No additional regularization (e.g., MixUp, CutMix) is applied, in order to isolate the architectural contribution of SVDA without confounding effects from advanced data augmentation. This ensures that any observed differences in performance or interpretability can be attributed directly to the attention mechanism.
\end{itemize}
Each image is divided into non-overlapping square patches. These patches are flattened and passed through a learned linear projection layer to produce fixed-length embeddings. Learnable positional encodings are added to the embedded sequence before input to the Transformer encoder.

\subsection{Model Architecture}

A consistent Vision Transformer (ViT) backbone is used across datasets, with architectural parameters scaled according to input resolution. For FashionMNIST, CIFAR-10, and CIFAR-100, the model uses an embedding dimension of 256, 4 Transformer layers, 4 attention heads per layer, an MLP hidden dimension of 1024, and a patch size of 4$\times$4. For ImageNet-100, the configuration is adjusted to an embedding dimension of 96, 8 layers, 2 attention heads, an MLP hidden dimension of 384, and a patch size of 8$\times$8.

A learnable [CLS] token is prepended to the patch sequence before being passed into the Transformer encoder. Each Transformer block applies Layer Normalization prior to both the self-attention mechanism and the feedforward sublayer. GELU is used as the activation function in all feedforward networks.

\subsection{Training and Optimization}

All models are trained for 100 epochs using the Adam optimizer. A learning rate of $3 \times 10^{-4}$ is used for FashionMNIST, CIFAR-10, and CIFAR-100, and $1 \times 10^{-4}$ for ImageNet-100. The batch size is set to 64 for all datasets except ImageNet-100, which uses a batch size of 32. Cross-entropy loss is used throughout, with no learning rate scheduling or early stopping applied. For SVDA, two lightweight regularization terms are applied:
\begin{itemize}
    \item Orthogonality regularization on the query/key projection matrices with weight $\lambda_{\text{ortho}} = 10^{-3}$ (or $3 \times 10^{-4}$ for ImageNet-100), to promote soft orthonormality.
    \item Spectral entropy regularization on $\Sigma$ with the same weight ($10^{-3}$), encouraging low-rank structure and concentration in attention spectra.
\end{itemize}
All experiments are repeated with fixed random seeds to ensure comparability, and training logs are saved for reproducibility.

\subsection{Evaluation Metrics and Logging}

Beyond classification accuracy and training times, we evaluate the interpretability and structure of attention using the six interpretability indicators introduced in~\cite{arampatzakis2025geometry}:
\begin{itemize}
  \item Spectral Entropy and Effective Rank to measure spectral complexity.
  \item Spectral Sparsity to show the proportion of negligible spectral weights in $\Sigma$.
  \item Angular Alignment as the average cosine similarity between normalized queries and keys.
  \item Selectivity Index as the degree of attention concentration across tokens.
  \item Perturbation Robustness as an index of change in attention maps under additive input noise.
\end{itemize}
All indicators are computed per head and per layer, logged at each training epoch. This enables multiple visualizations, including time-evolving characterization.

\section{Results}
\newcommand{\imgswidth}{0.23\linewidth}
\newcommand{\labelswidth}{0.05\linewidth}

We evaluated the interpretability and structural behavior of SVDA-based Vision Transformers using the six attention indicators across all layers and heads. Each indicator provides insight into a distinct aspect of the attention mechanism's structure, stability, and informativeness.

\subsection{Accuracy and Training Time Comparison}

\begin{figure*}[!t]
\centering
\begin{minipage}{\imgswidth} \centering \textbf{FashionMNIST} \end{minipage}
\hfill
\begin{minipage}{\imgswidth} \centering \textbf{CIFAR-10} \end{minipage}
\hfill
\begin{minipage}{\imgswidth} \centering \textbf{CIFAR-100} \end{minipage}
\hfill
\begin{minipage}{\imgswidth} \centering \textbf{ImageNet-100} \end{minipage}
\par\vspace{1em}
\begin{minipage}{\imgswidth} \centering \includegraphics[width=\linewidth]{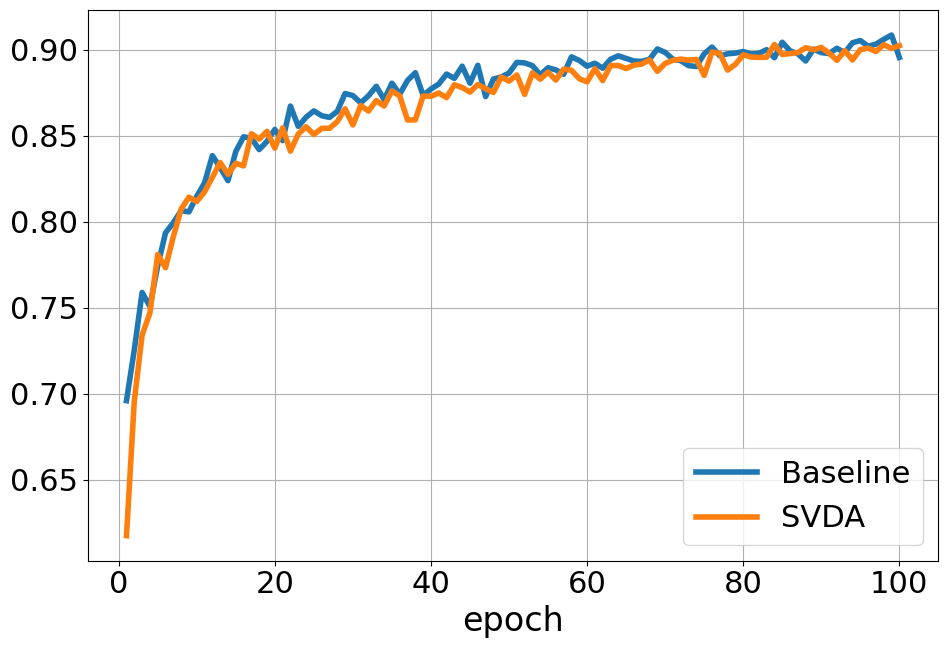} \end{minipage}
\hfill
\begin{minipage}{\imgswidth} \centering \includegraphics[width=\linewidth]{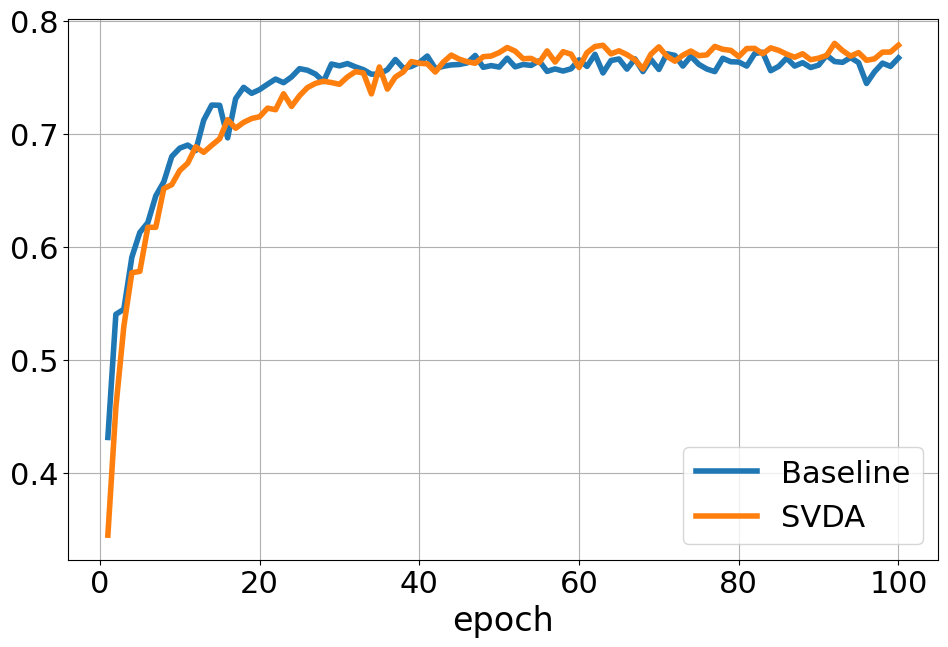} \end{minipage}
\hfill
\begin{minipage}{\imgswidth} \centering \includegraphics[width=\linewidth]{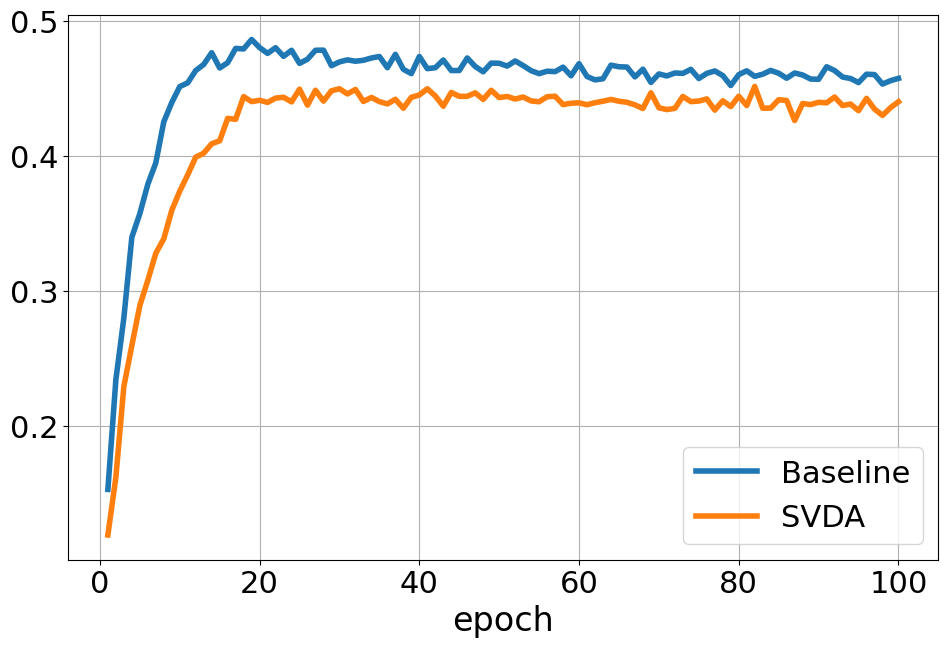} \end{minipage}
\hfill
\begin{minipage}{\imgswidth} \centering \includegraphics[width=\linewidth]{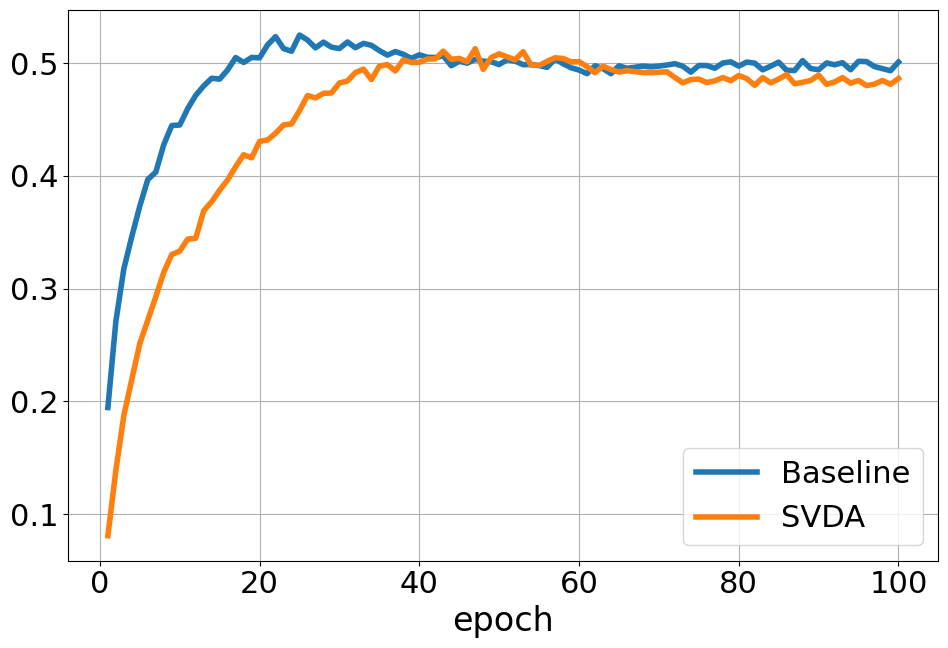} \end{minipage}
\par\vspace{0.5em}
\caption{Accuracy over epochs for baseline and SVDA models across all datasets. The two models show nearly identical learning trajectories.}
\label{fig:svda_accuracy}
\end{figure*}

\begin{figure*}[!t]
\centering
\begin{minipage}{\imgswidth} \centering \textbf{FashionMNIST} \end{minipage}
\hfill
\begin{minipage}{\imgswidth} \centering \textbf{CIFAR-10} \end{minipage}
\hfill
\begin{minipage}{\imgswidth} \centering \textbf{CIFAR-100} \end{minipage}
\hfill
\begin{minipage}{\imgswidth} \centering \textbf{ImageNet-100} \end{minipage}
\par\vspace{1em}
\begin{minipage}{\imgswidth} \centering \includegraphics[width=\linewidth]{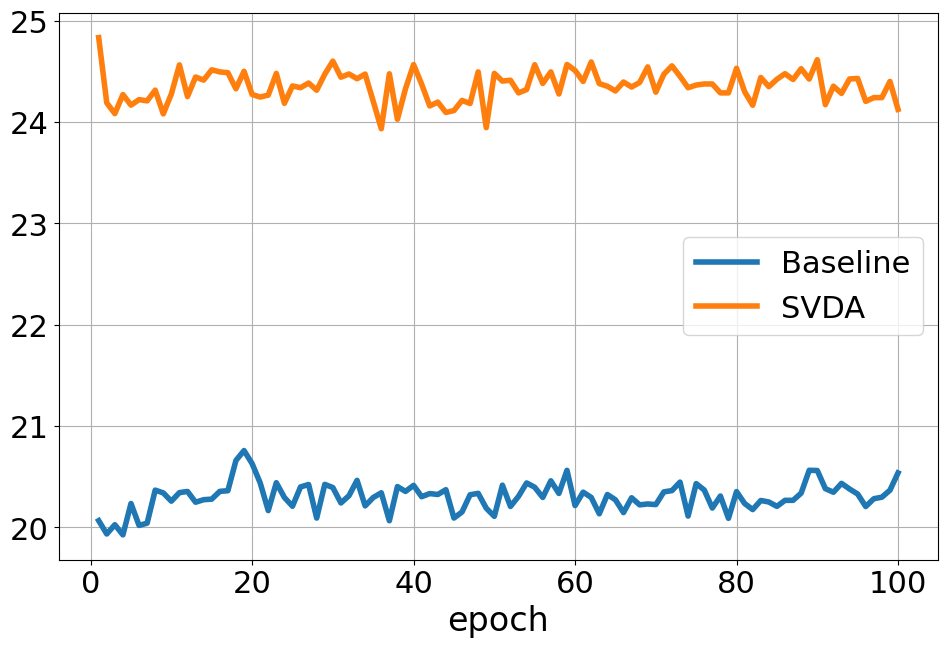} \end{minipage}
\hfill
\begin{minipage}{\imgswidth} \centering \includegraphics[width=\linewidth]{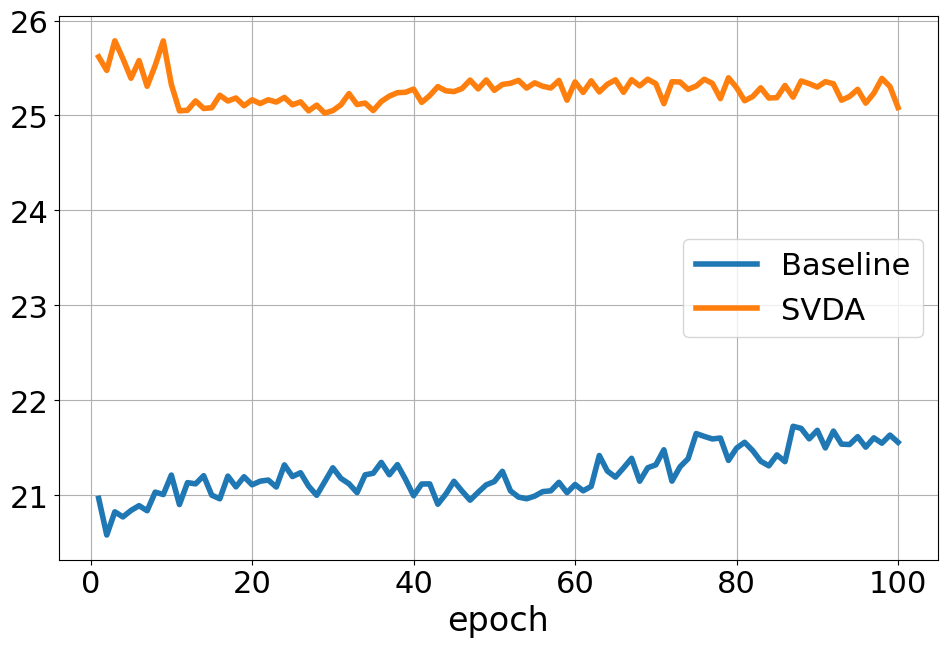} \end{minipage}
\hfill
\begin{minipage}{\imgswidth} \centering \includegraphics[width=\linewidth]{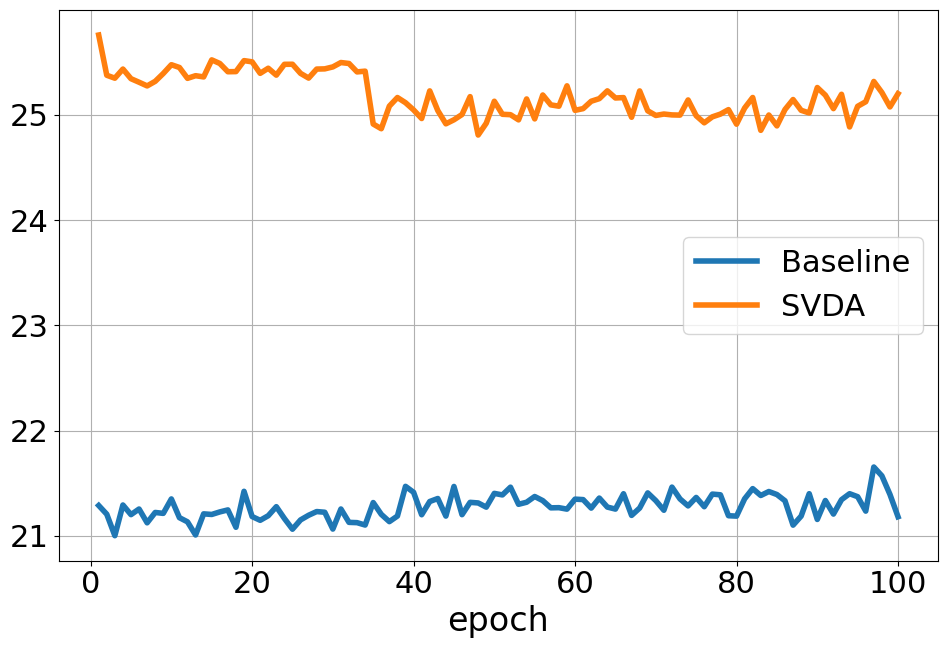} \end{minipage}
\hfill
\begin{minipage}{\imgswidth} \centering \includegraphics[width=\linewidth]{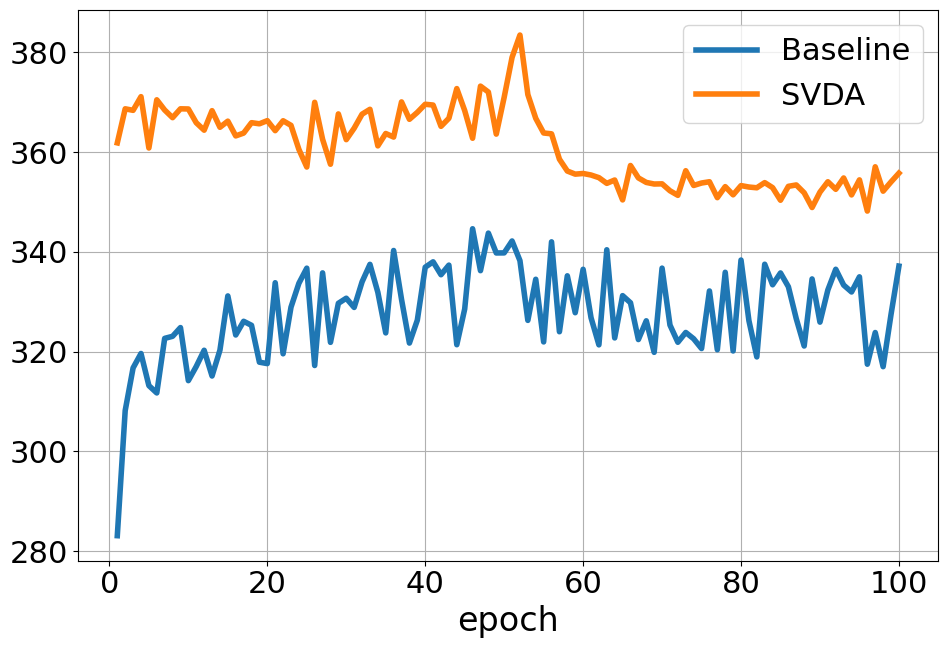} \end{minipage}
\par\vspace{0.5em}
\caption{Training time per epoch for baseline and SVDA models. SVDA introduces an average overhead of approximately 17\% due to its additional spectral normalization and modulation steps.}
\label{fig:svda_training_time}
\end{figure*}

To assess the computational implications of SVDA, we compared its accuracy and runtime against the baseline Vision Transformer across all datasets. 

\figurename~\ref{fig:svda_accuracy} confirms that classification accuracy evolves nearly identically for both models throughout training. While both compared architectures maintain comparative and high validation accuracy on simpler datasets, such as FashionMNIST and CIFAR-10, we observe a pronounced generalization gap in CIFAR-100 and ImageNet-100, where training accuracy exceeds 90\% but validation accuracy plateaus below 50\%. This overfitting behavior can be attributed to the limited capacity of the experimental architectures used---4 transformer layers with 4 heads (or 8 layers with 2 heads)---which are significantly shallower than standard Vision Transformers designed for high-resolution, high-diversity datasets. For example, ViT-Base employs 12 layers and 12 heads with a hidden size of 768 and is typically pretrained on massive datasets like JFT-300M before fine-tuning on ImageNet~\cite{dosovitskiy2020image}. In contrast, our models are trained from scratch on modest-size datasets, which have been shown to be critical for generalization in transformer-based vision models~\cite{touvron2021training, xiao2021early}. Furthermore, transformers are known to require substantial depth and width to disentangle fine-grained classes effectively, particularly in datasets like CIFAR-100 and ImageNet, which exhibit high intra-class variation and inter-class similarity~\cite{steiner2021train}. These findings highlight both the limitations of compact transformer architectures and the need for capacity-aware interpretability analysis, as SVDA continues to reveal distinct structural properties even under suboptimal generalization.

\figurename~\ref{fig:svda_training_time} provides insight on the training time per epoch, showing a consistently higher trend for SVDA. This overhead stems primarily from the additional matrix normalization and spectral modulation operations introduced in the attention layers. Notably, the interpretability benefits of SVDA are achieved without compromising predictive performance, albeit with a modest computational cost. While SVDA introduces only minimal architectural overhead, adding fewer than 0.04\% more trainable parameters in all cases, it consistently incurs a modest training time slowdown across datasets. On FashionMNIST, CIFAR-10, and CIFAR-100, SVDA-trained models exhibit a relative slowdown of approximately $18-20\%$, while on the more computationally intensive ImageNet-100 task, the slowdown drops to $\sim10\%$. This moderate increase in wall-clock time occurs despite the fact that SVDA models consistently report lower Multiply-Accumulate Operations (MACs) than their baseline counterparts. For instance, on ImageNet-100, the SVDA variant computes 1.68 GMACs per forward pass versus 2.63 GMACs for the baseline, a 36\% reduction. This paradox, lower theoretical computation yet slower real-world training, is likely attributable to implementation-level inefficiencies. SVDA involves SVD-based projection steps and orthogonality constraints, which are not as highly optimized for GPU parallelization as standard matrix multiplications. In contrast, baseline attention implementations benefit from mature, low-level CUDA kernels and fused operations available in modern deep learning frameworks such as PyTorch, in which we relied. Recent studies on efficient attention variants (e.g.,~\cite{dao2022flashattention}) emphasize that raw FLOP or MAC counts do not always translate linearly to performance on hardware accelerators, particularly when kernel fusion, memory layout, and parallelization are involved.

Thus, while SVDA incurs an average 17.9\% training slowdown across datasets, this cost is relatively modest given the structural interpretability and spectral regularization benefits it provides. Future versions may reduce this gap through custom kernel optimization and batching strategies that better exploit GPU hardware.

\subsection{Metric Evolution and Layerwise Analysis}

\begin{figure*}[!htbp]
\centering

\begin{minipage}{\labelswidth} \vphantom{\textbf{a}} \end{minipage}
\hfill
\begin{minipage}{\imgswidth} \centering \textbf{FashionMNIST} \end{minipage}
\hfill
\begin{minipage}{\imgswidth} \centering \textbf{CIFAR-10} \end{minipage}
\hfill
\begin{minipage}{\imgswidth} \centering \textbf{CIFAR-100} \end{minipage}
\hfill
\begin{minipage}{\imgswidth} \centering \textbf{ImageNet-100} \end{minipage}
\par\vspace{1em}

\begin{minipage}{\labelswidth} \centering \textbf{a} \end{minipage}
\hfill
\begin{minipage}{\imgswidth} \centering \includegraphics[width=\linewidth]{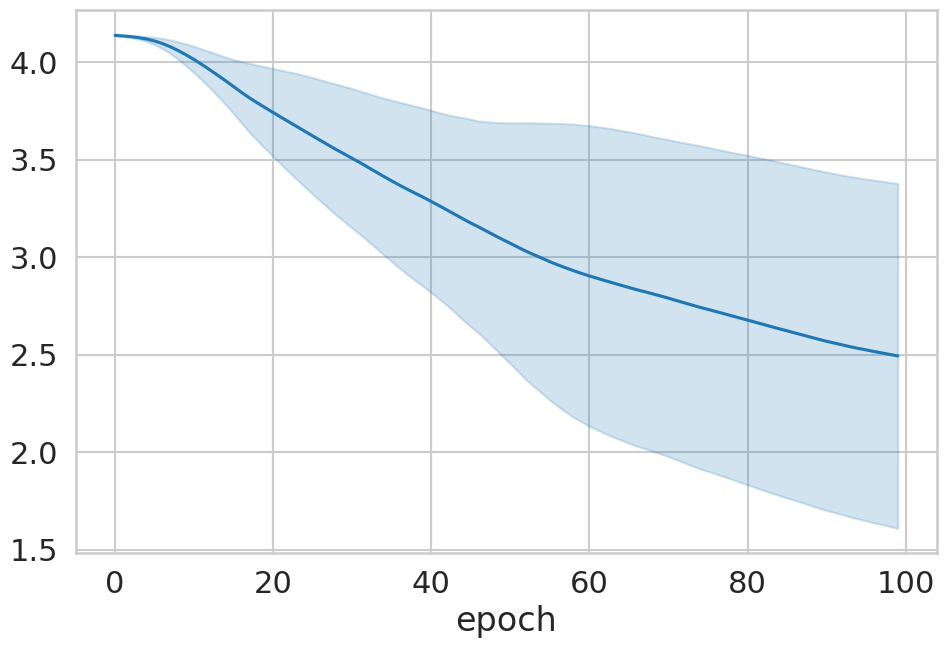} \end{minipage}
\hfill
\begin{minipage}{\imgswidth} \centering \includegraphics[width=\linewidth]{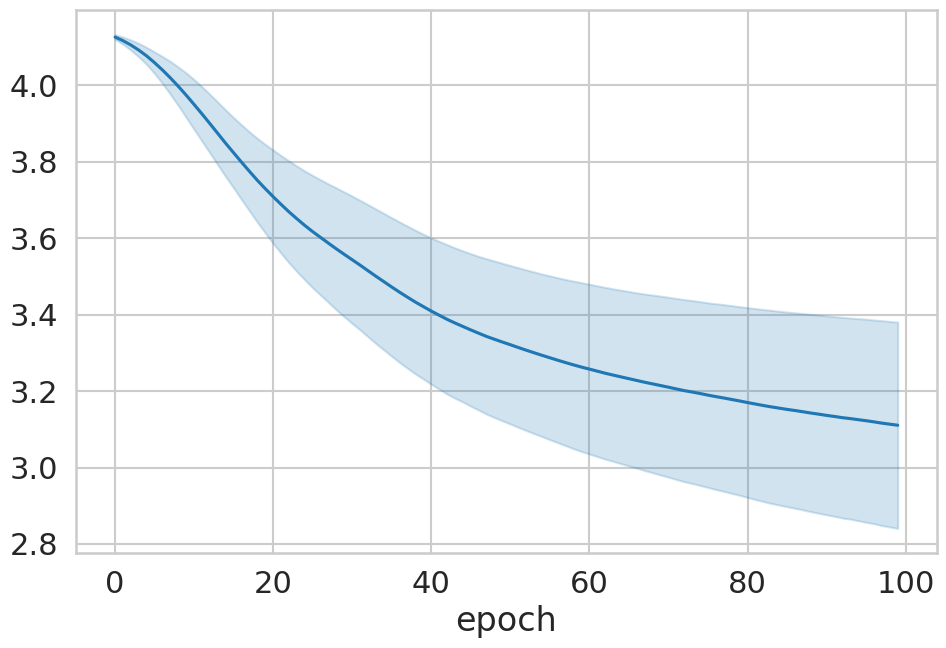} \end{minipage}
\hfill
\begin{minipage}{\imgswidth} \centering \includegraphics[width=\linewidth]{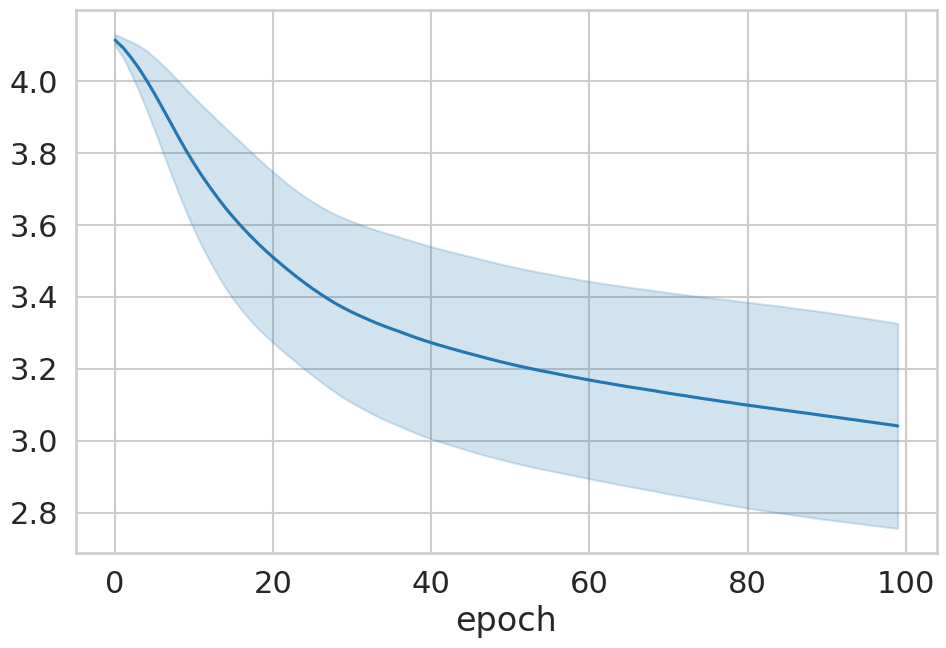} \end{minipage}
\hfill
\begin{minipage}{\imgswidth} \centering \includegraphics[width=\linewidth]{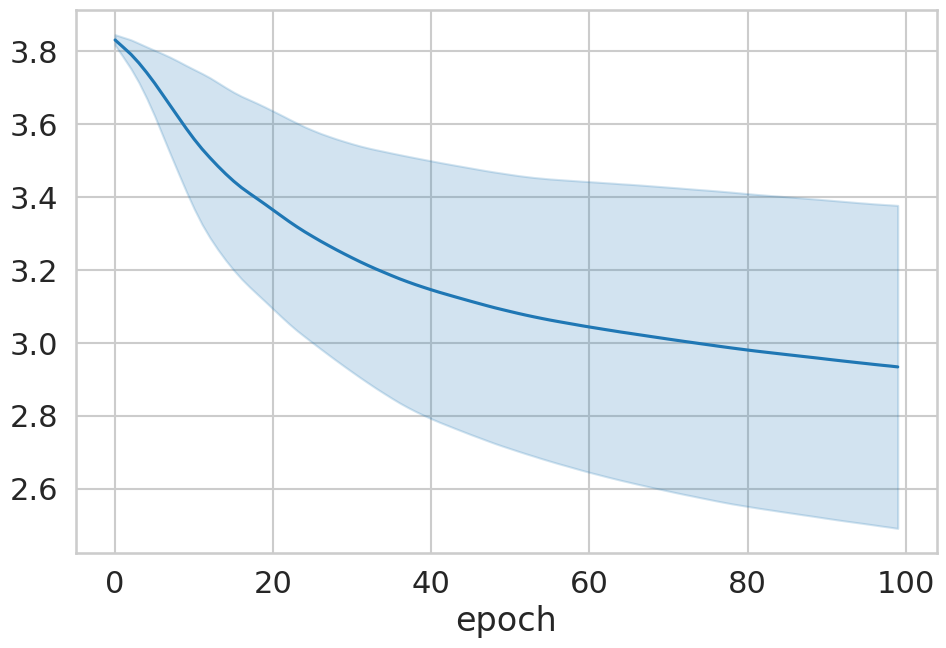} \end{minipage}
\par\vspace{2.2em}

\begin{minipage}{\labelswidth} \centering \textbf{b} \end{minipage}
\hfill
\begin{minipage}{\imgswidth} \centering \includegraphics[width=\linewidth]{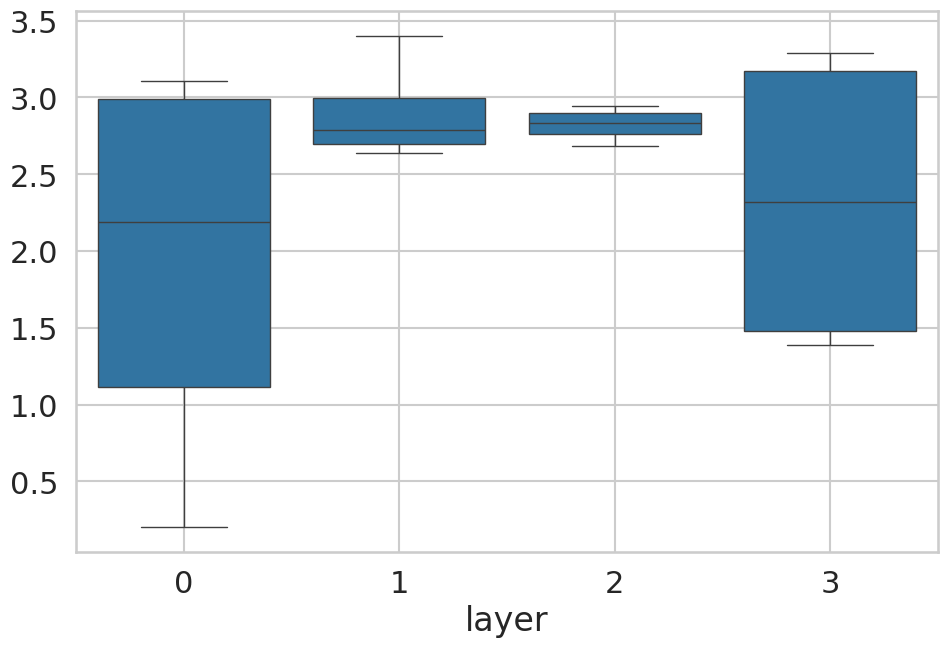} \end{minipage}
\hfill
\begin{minipage}{\imgswidth} \centering \includegraphics[width=\linewidth]{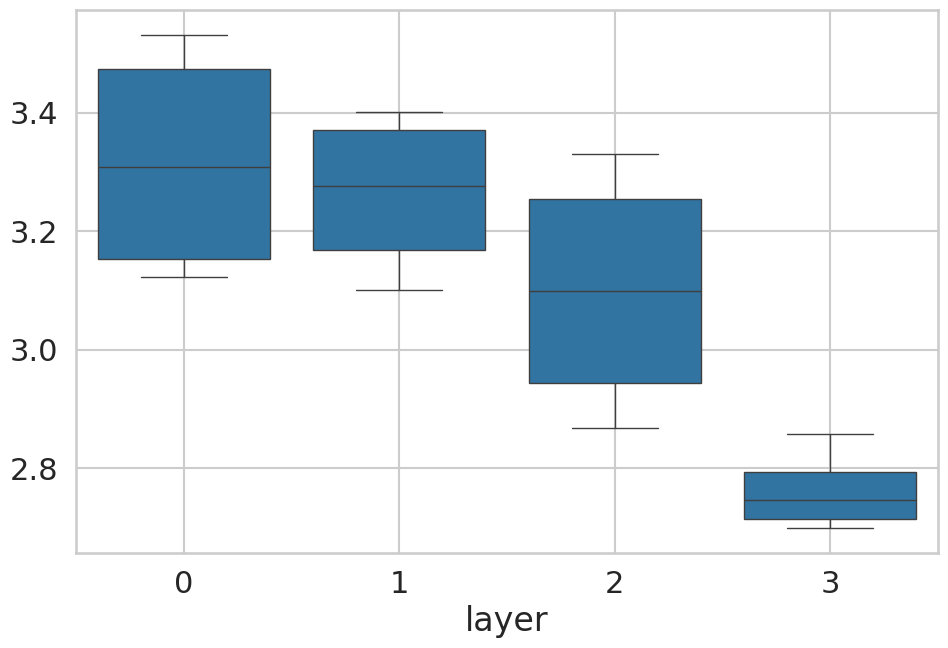} \end{minipage}
\hfill
\begin{minipage}{\imgswidth} \centering \includegraphics[width=\linewidth]{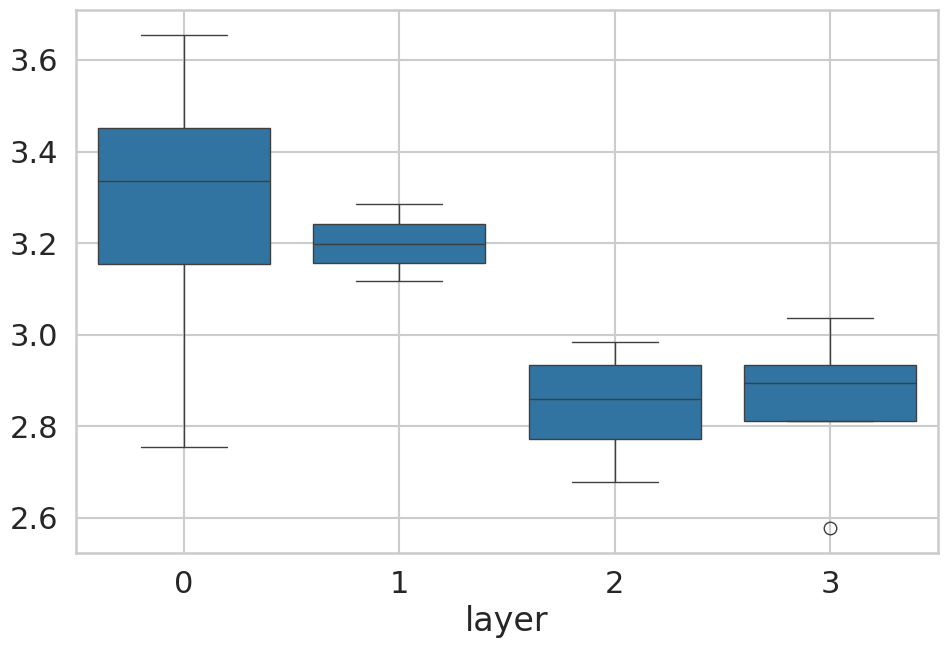} \end{minipage}
\hfill
\begin{minipage}{\imgswidth} \centering \includegraphics[width=\linewidth]{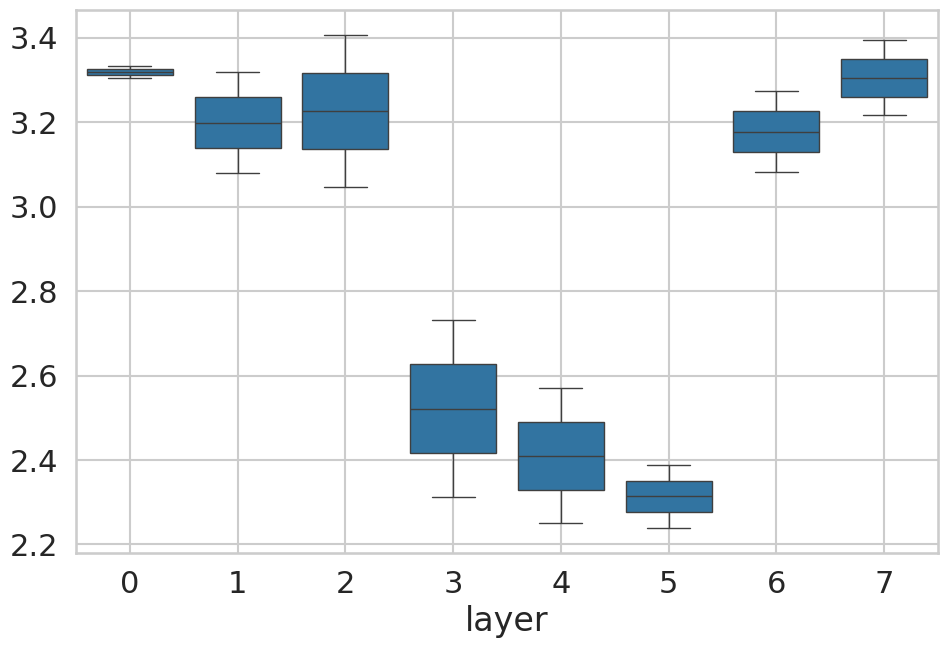} \end{minipage}
\par\vspace{2.2em}

\begin{minipage}{\labelswidth} \centering \textbf{c} \end{minipage}
\hfill
\begin{minipage}{\imgswidth} \centering \includegraphics[width=\linewidth]{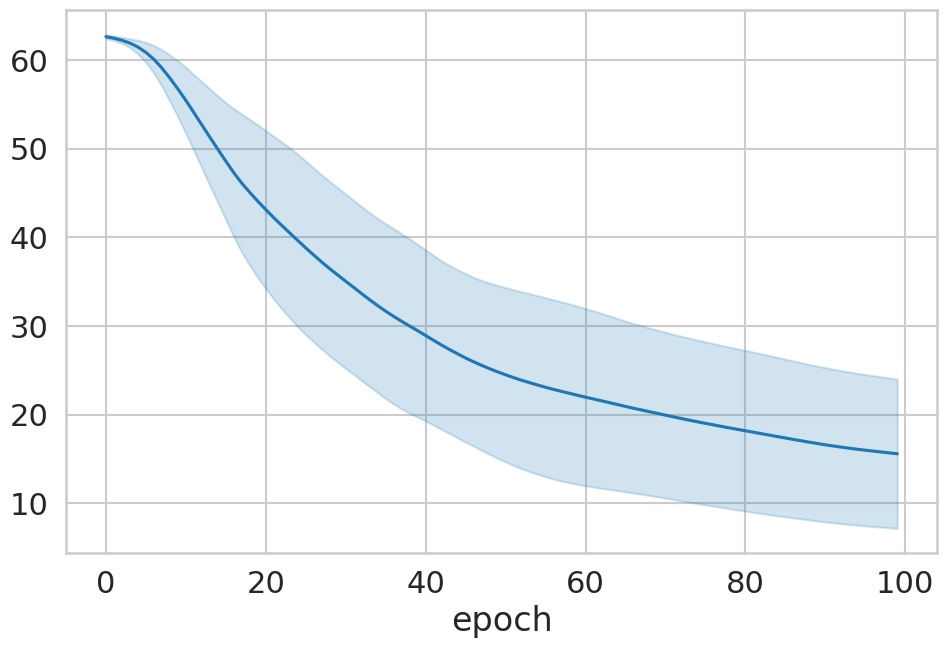} \end{minipage}
\hfill
\begin{minipage}{\imgswidth} \centering \includegraphics[width=\linewidth]{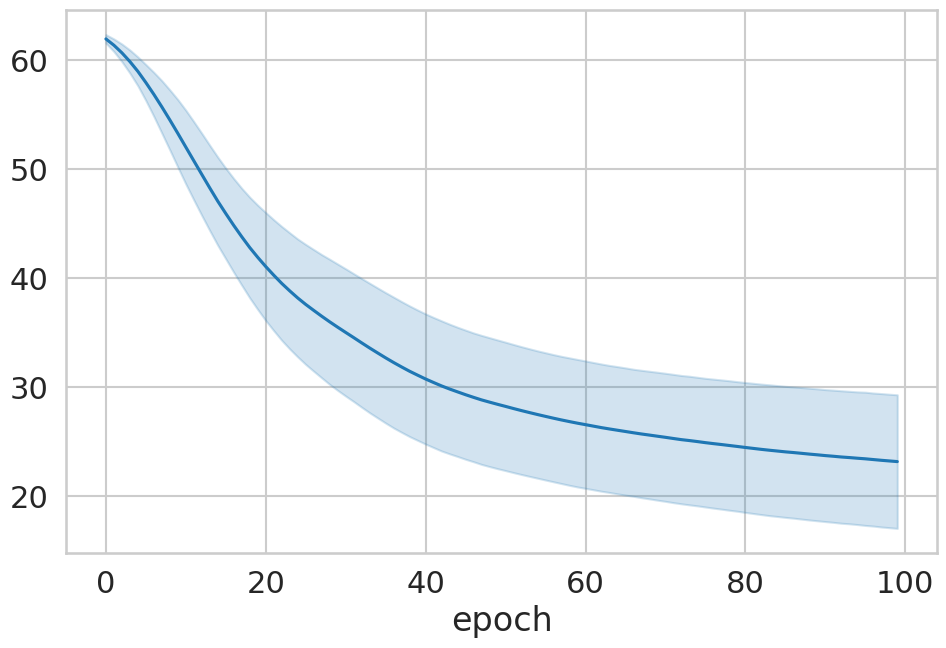} \end{minipage}
\hfill
\begin{minipage}{\imgswidth} \centering \includegraphics[width=\linewidth]{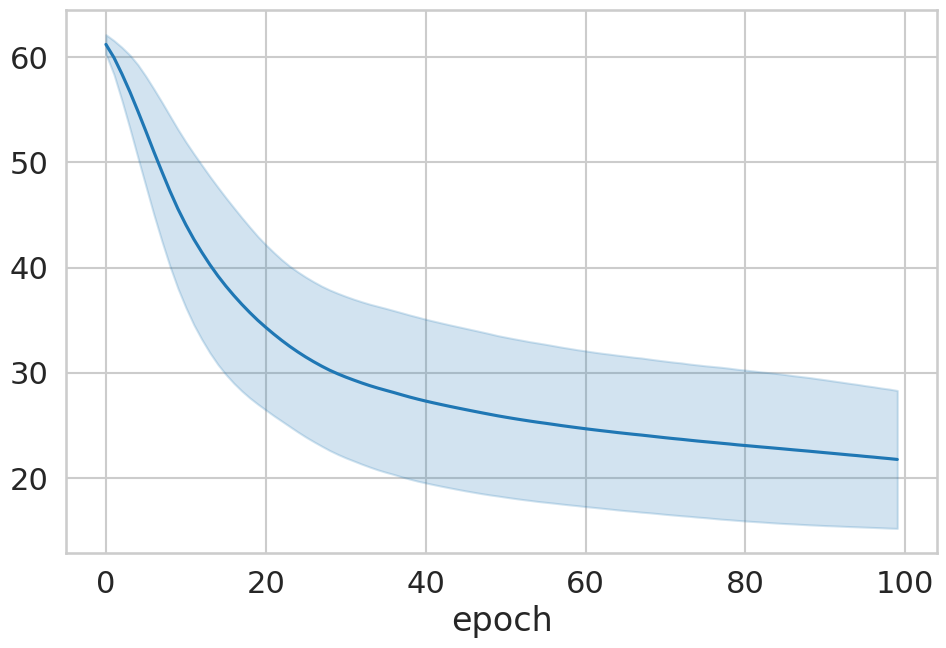} \end{minipage}
\hfill
\begin{minipage}{\imgswidth} \centering \includegraphics[width=\linewidth]{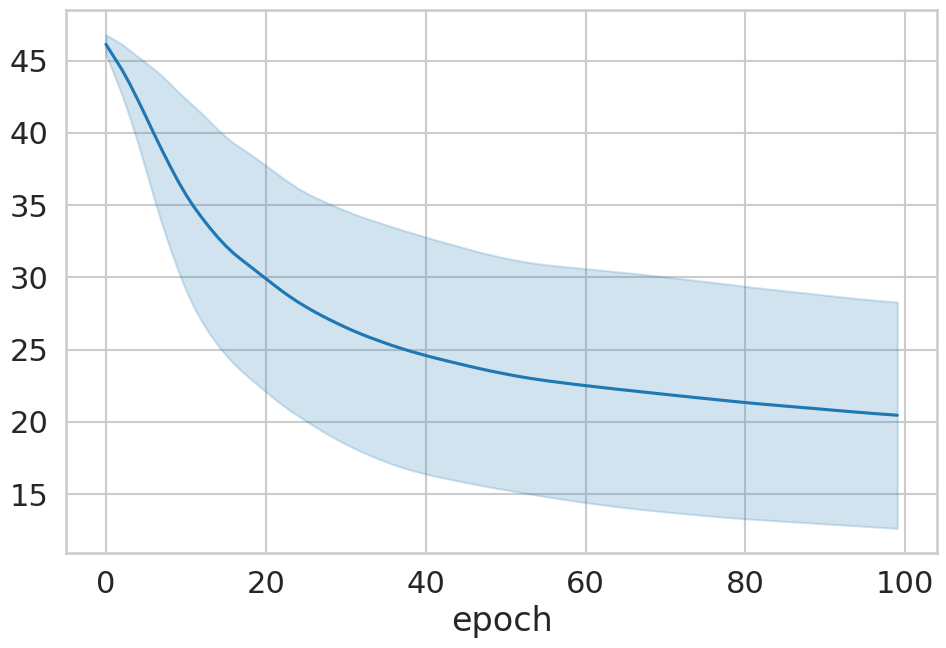} \end{minipage}
\par\vspace{2.2em}

\begin{minipage}{\labelswidth} \centering \textbf{d} \end{minipage}
\hfill
\begin{minipage}{\imgswidth} \centering \includegraphics[width=\linewidth]{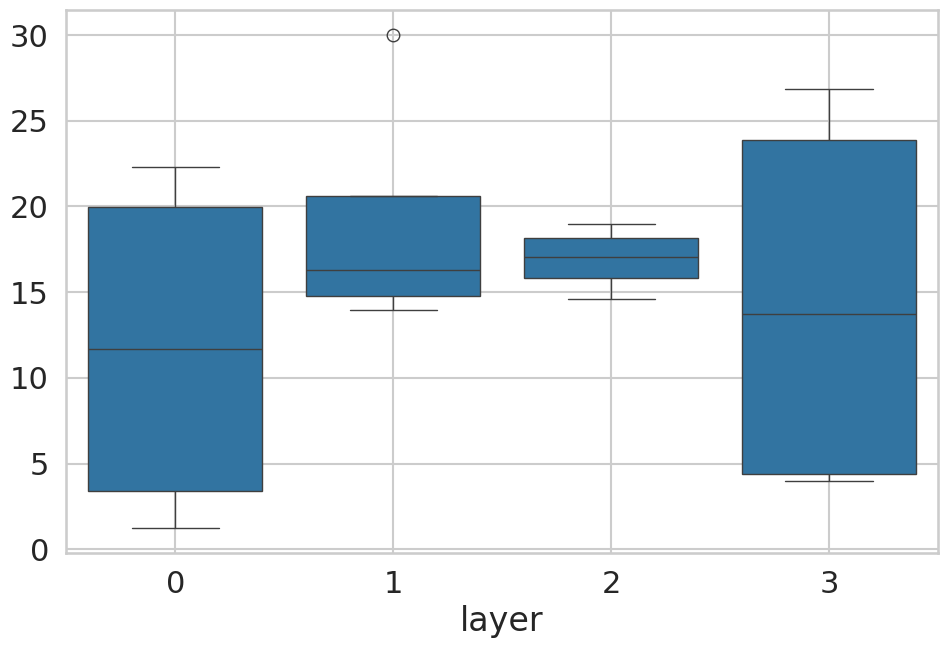} \end{minipage}
\hfill
\begin{minipage}{\imgswidth} \centering \includegraphics[width=\linewidth]{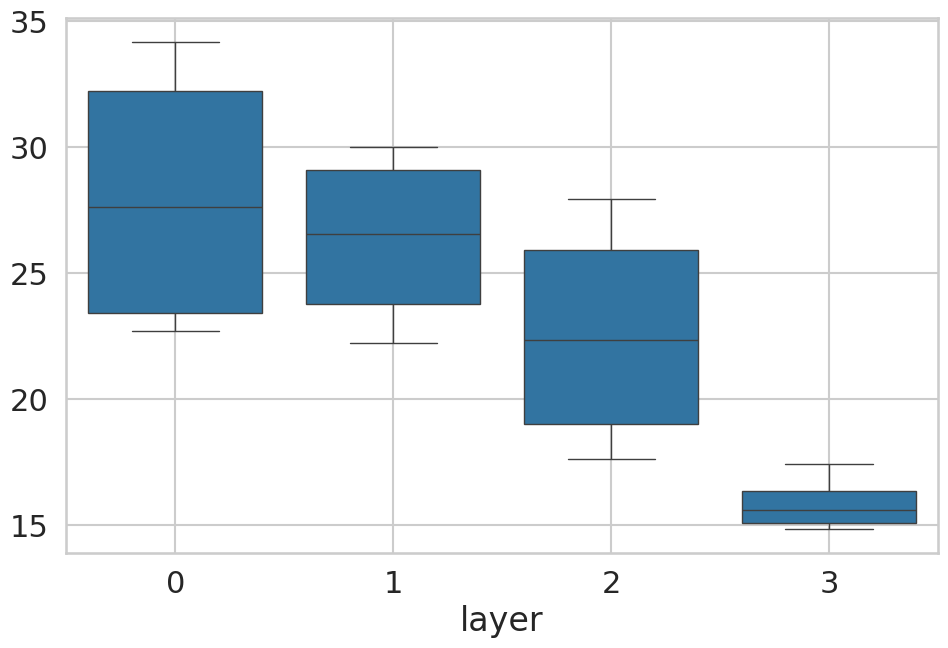} \end{minipage}
\hfill
\begin{minipage}{\imgswidth} \centering \includegraphics[width=\linewidth]{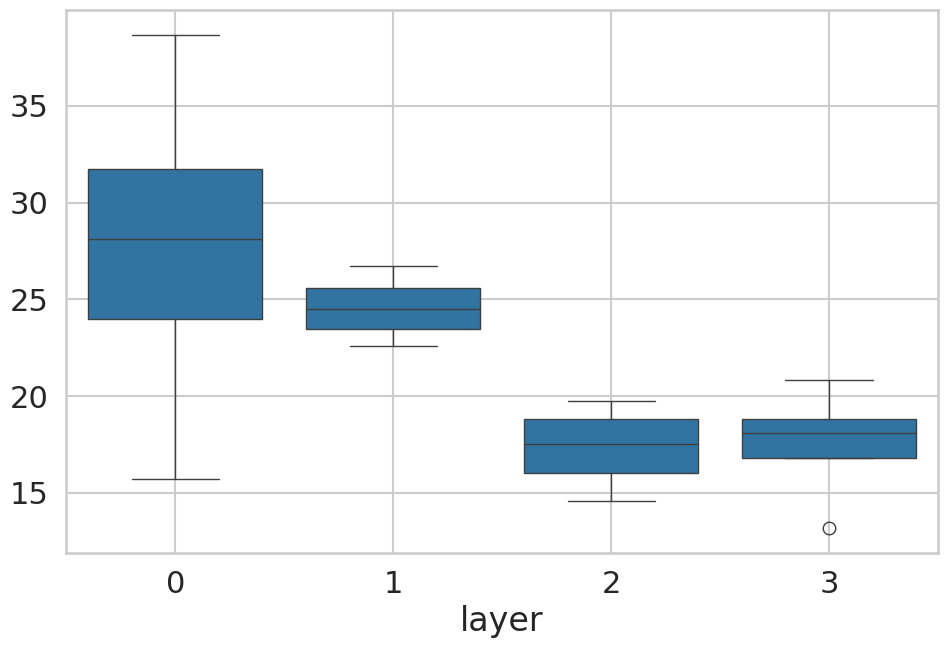} \end{minipage}
\hfill
\begin{minipage}{\imgswidth} \centering \includegraphics[width=\linewidth]{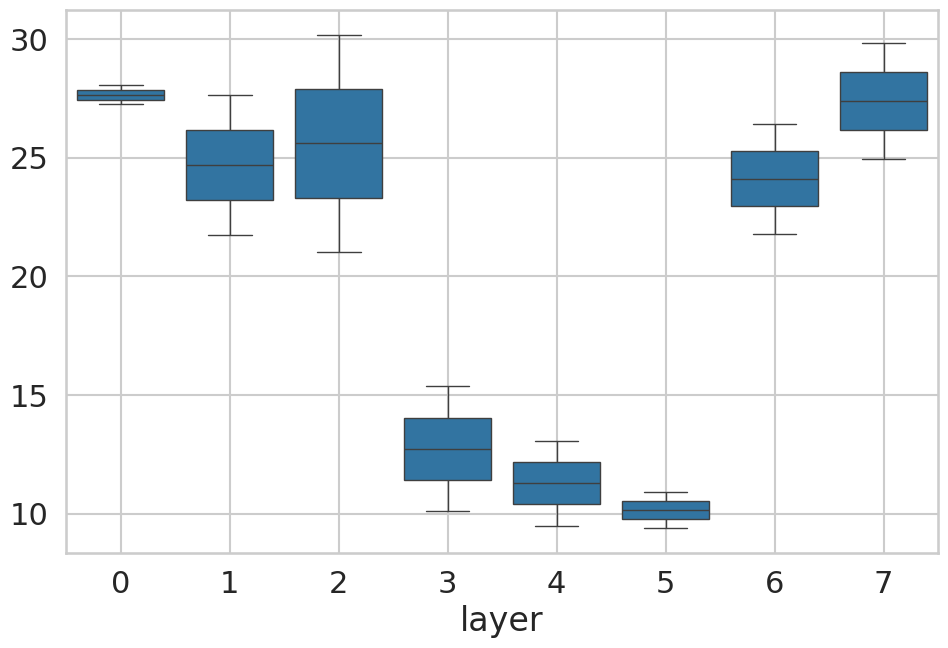} \end{minipage}
\par\vspace{2.2em}

\begin{minipage}{\labelswidth} \centering \textbf{e} \end{minipage}
\hfill
\begin{minipage}{\imgswidth} \centering \includegraphics[width=\linewidth]{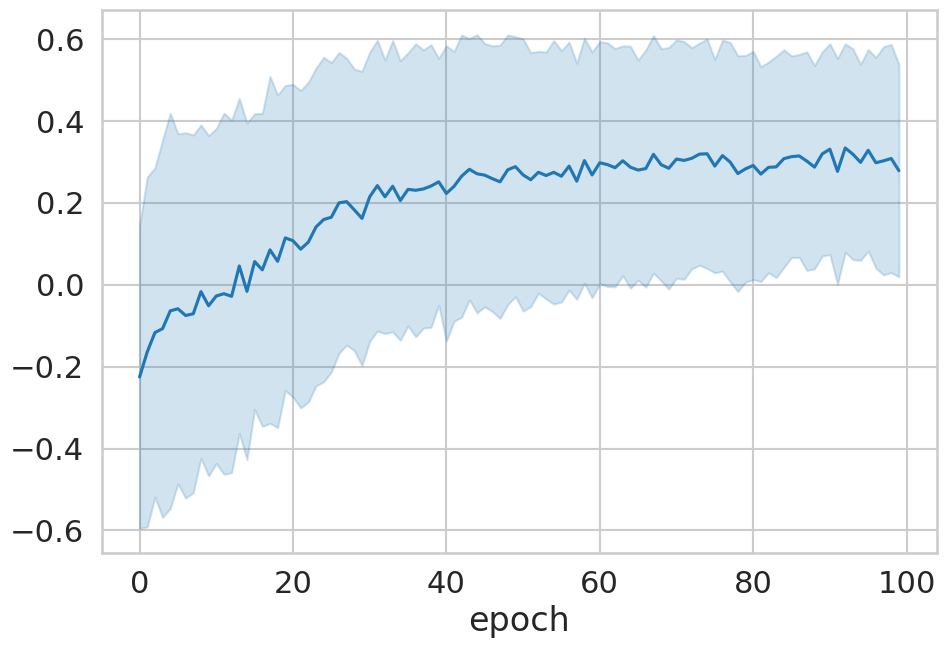} \end{minipage}
\hfill
\begin{minipage}{\imgswidth} \centering \includegraphics[width=\linewidth]{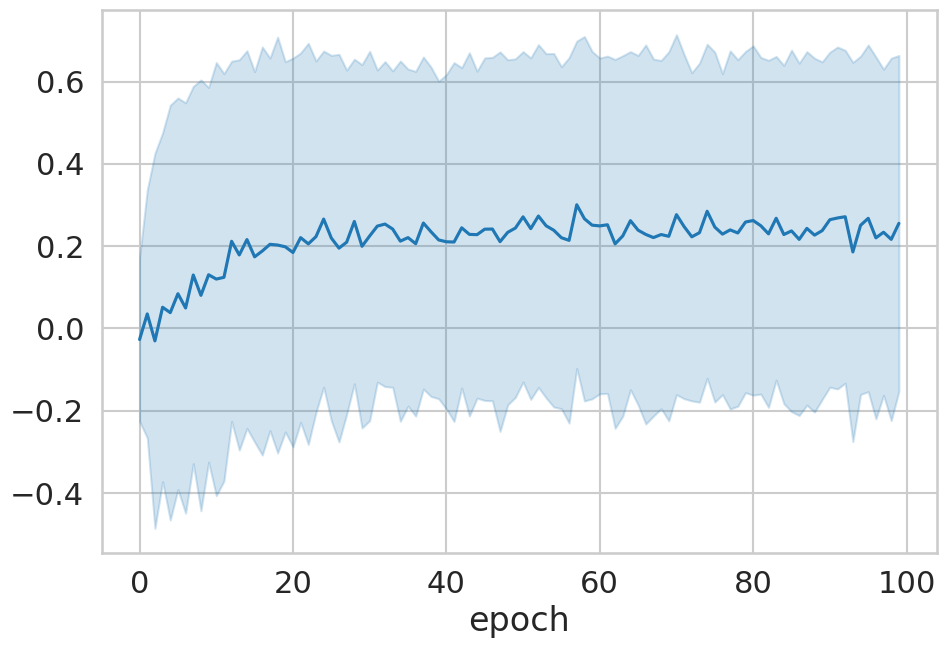} \end{minipage}
\hfill
\begin{minipage}{\imgswidth} \centering \includegraphics[width=\linewidth]{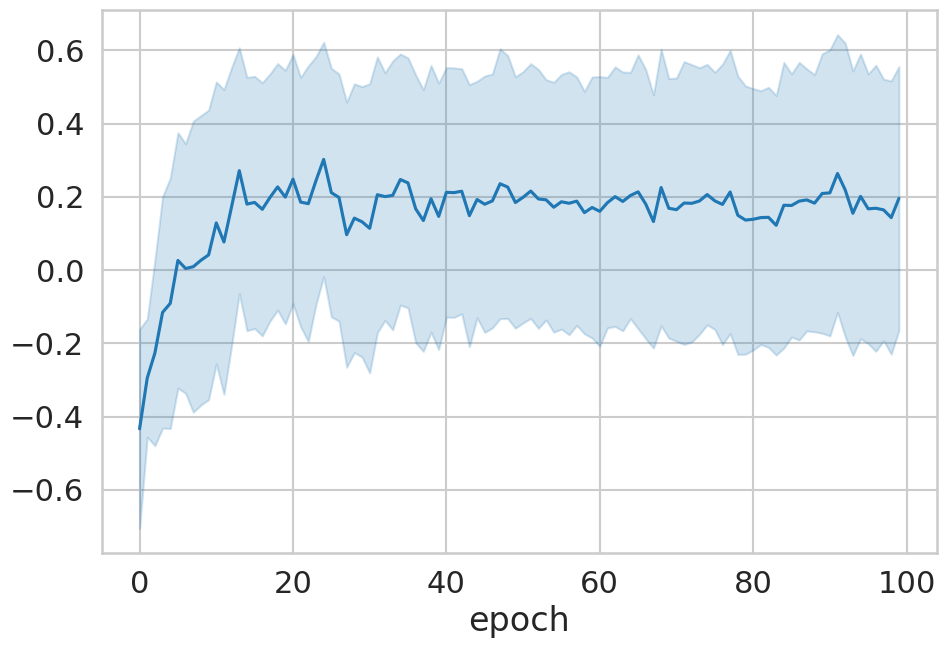} \end{minipage}
\hfill
\begin{minipage}{\imgswidth} \centering \includegraphics[width=\linewidth]{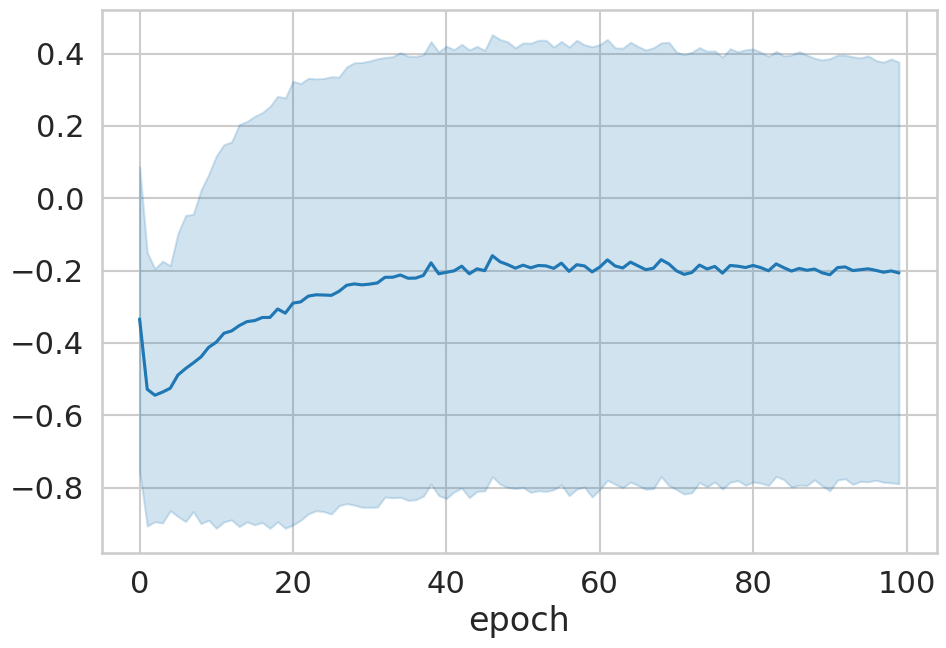} \end{minipage}
\par\vspace{2.2em}

\begin{minipage}{\labelswidth} \centering \textbf{f} \end{minipage}
\hfill
\begin{minipage}{\imgswidth} \centering \includegraphics[width=\linewidth]{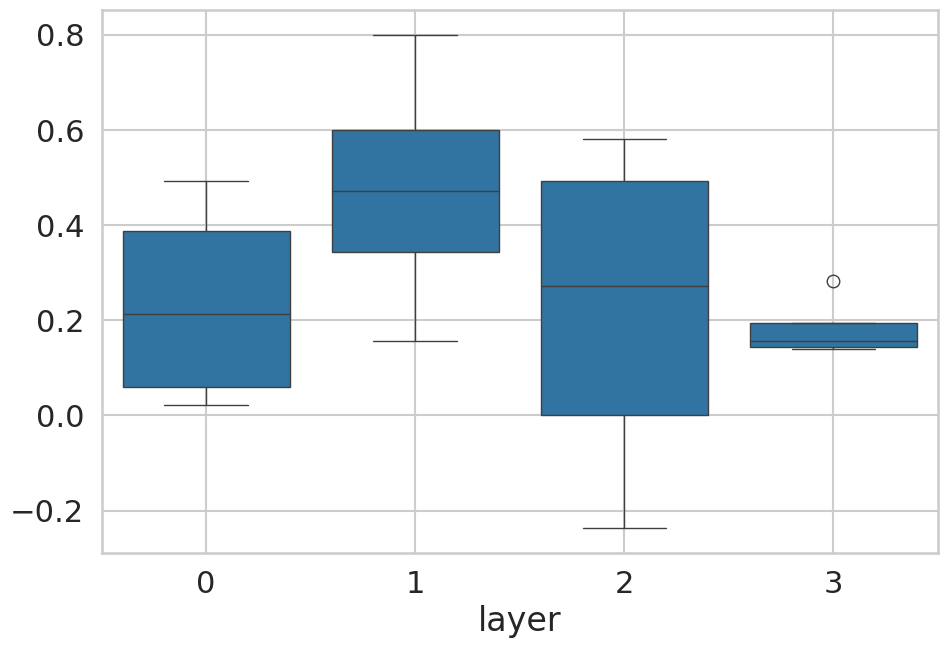} \end{minipage}
\hfill
\begin{minipage}{\imgswidth} \centering \includegraphics[width=\linewidth]{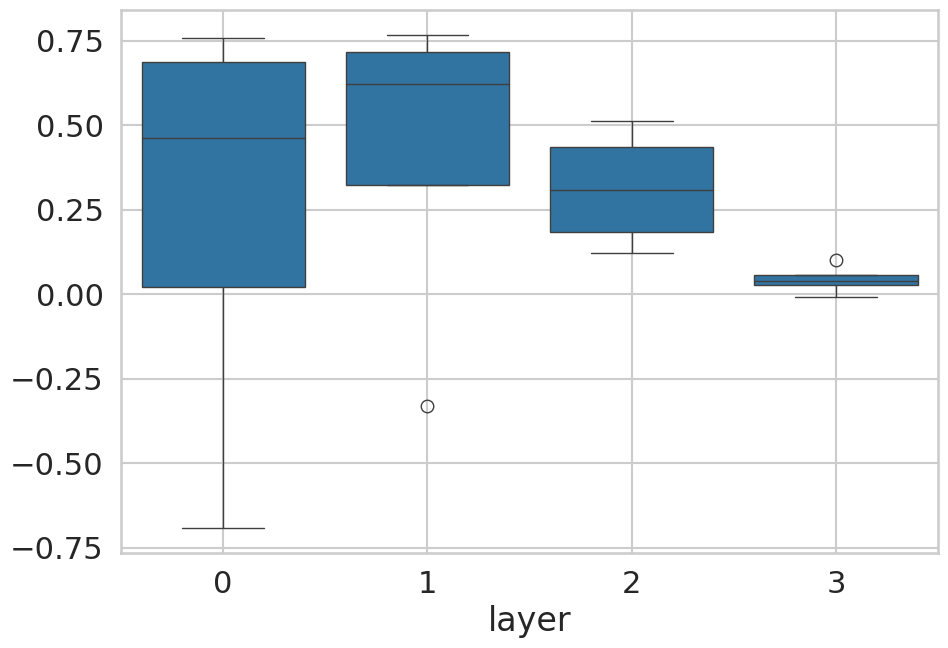} \end{minipage}
\hfill
\begin{minipage}{\imgswidth} \centering \includegraphics[width=\linewidth]{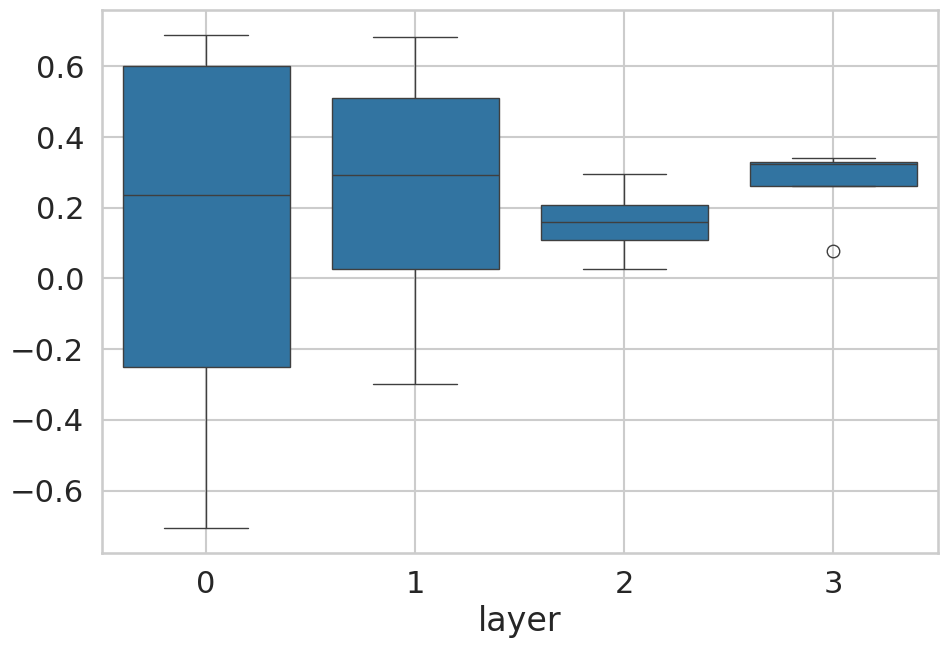} \end{minipage}
\hfill
\begin{minipage}{\imgswidth} \centering \includegraphics[width=\linewidth]{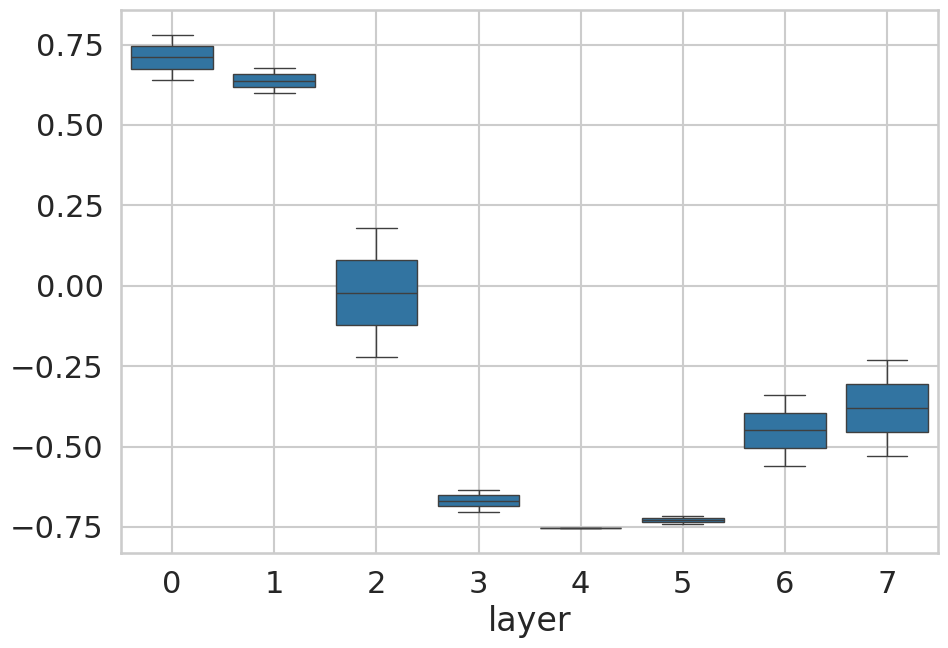} \end{minipage}
\par\vspace{0.5em}

\caption{SVDA interpretability and structural attention diagnostics across datasets and interpretability indicators (part 1 of 2): (a) Spectral Entropy evolution per epoch; (b) Spectral Entropy per layer; (c) Effective Rank evolution per epoch; (d) Effective Rank per layer; (e) Angular Alignment evolution per epoch; (f) Angular Alignment per layer.}
\label{fig:svda_grid_a}
\end{figure*}

\begin{figure*}[!htbp]
\centering

\begin{minipage}{\labelswidth} \vphantom{\textbf{a}} \end{minipage}
\hfill
\begin{minipage}{\imgswidth} \centering \textbf{FashionMNIST} \end{minipage}
\hfill
\begin{minipage}{\imgswidth} \centering \textbf{CIFAR-10} \end{minipage}
\hfill
\begin{minipage}{\imgswidth} \centering \textbf{CIFAR-100} \end{minipage}
\hfill
\begin{minipage}{\imgswidth} \centering \textbf{ImageNet-100} \end{minipage}
\par\vspace{1.5em}

\begin{minipage}{\labelswidth} \centering \textbf{g} \end{minipage}
\hfill
\begin{minipage}{\imgswidth} \centering \includegraphics[width=\linewidth]{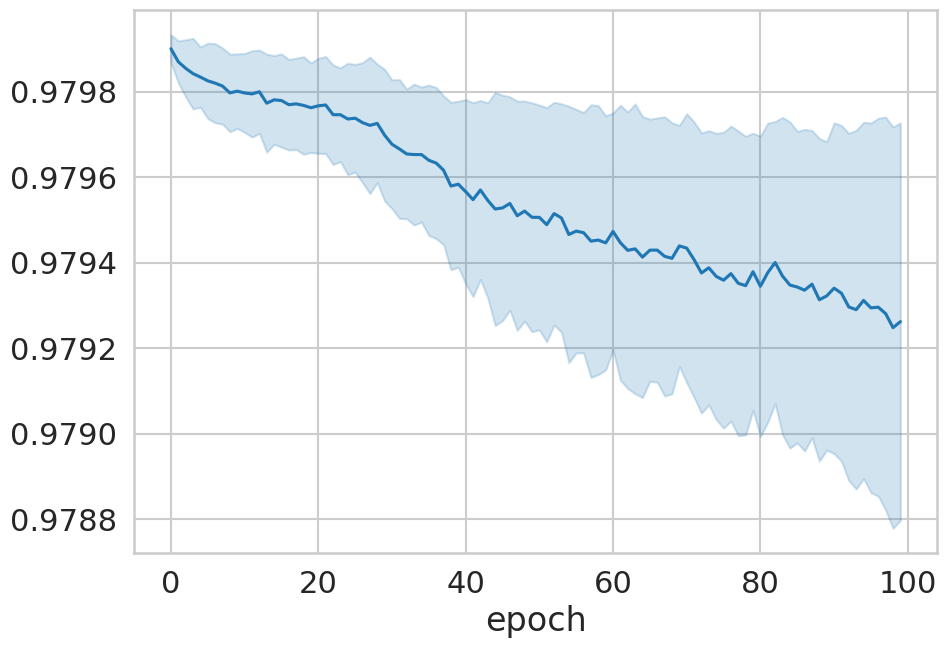} \end{minipage}
\hfill
\begin{minipage}{\imgswidth} \centering \includegraphics[width=\linewidth]{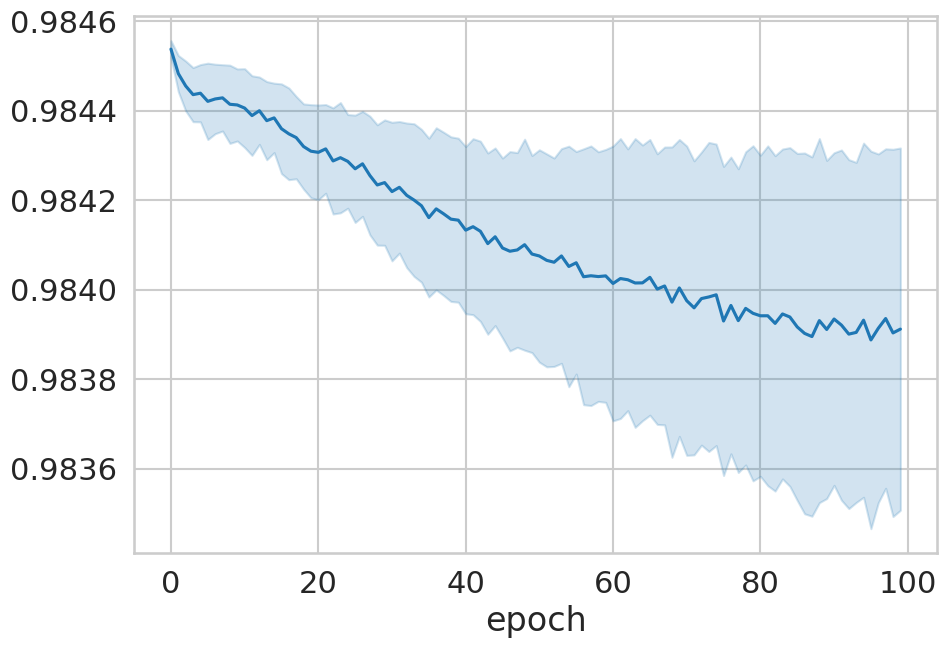} \end{minipage}
\hfill
\begin{minipage}{\imgswidth} \centering \includegraphics[width=\linewidth]{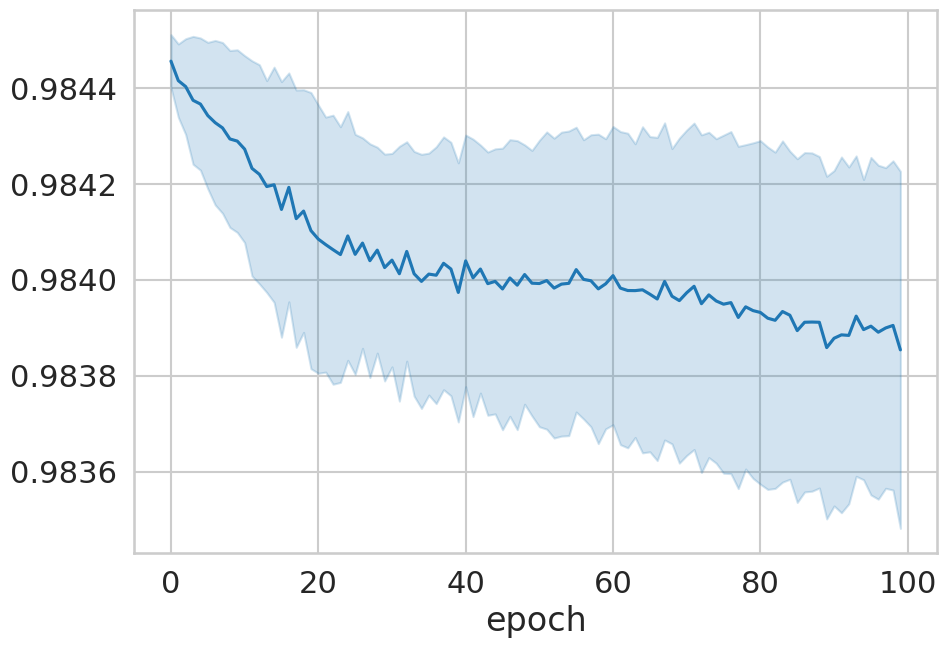} \end{minipage}
\hfill
\begin{minipage}{\imgswidth} \centering \includegraphics[width=\linewidth]{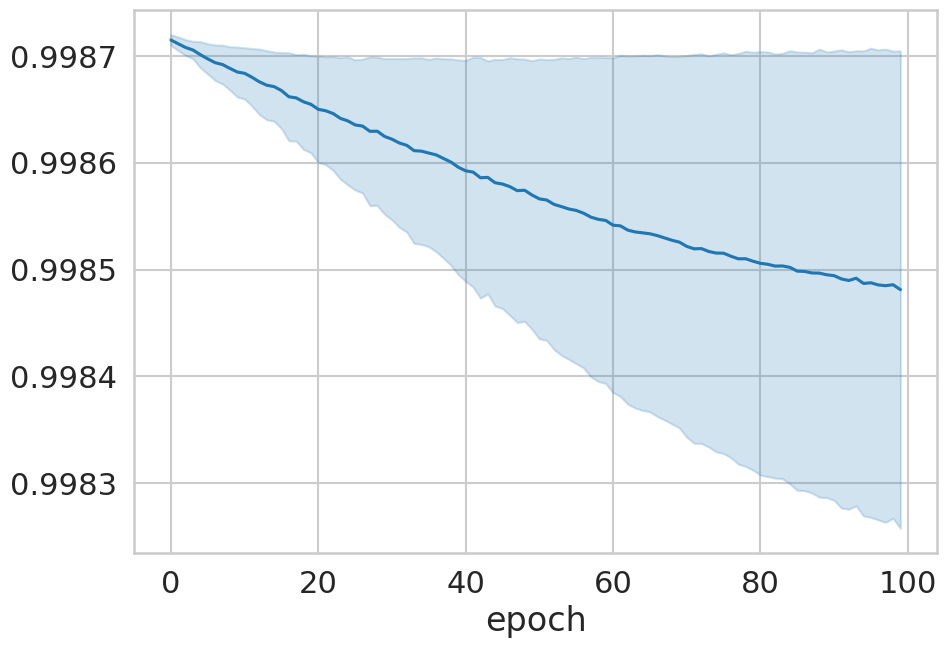} \end{minipage}
\par\vspace{2.2em}

\begin{minipage}{\labelswidth} \centering \textbf{h} \end{minipage}
\hfill
\begin{minipage}{\imgswidth} \centering \includegraphics[width=\linewidth]{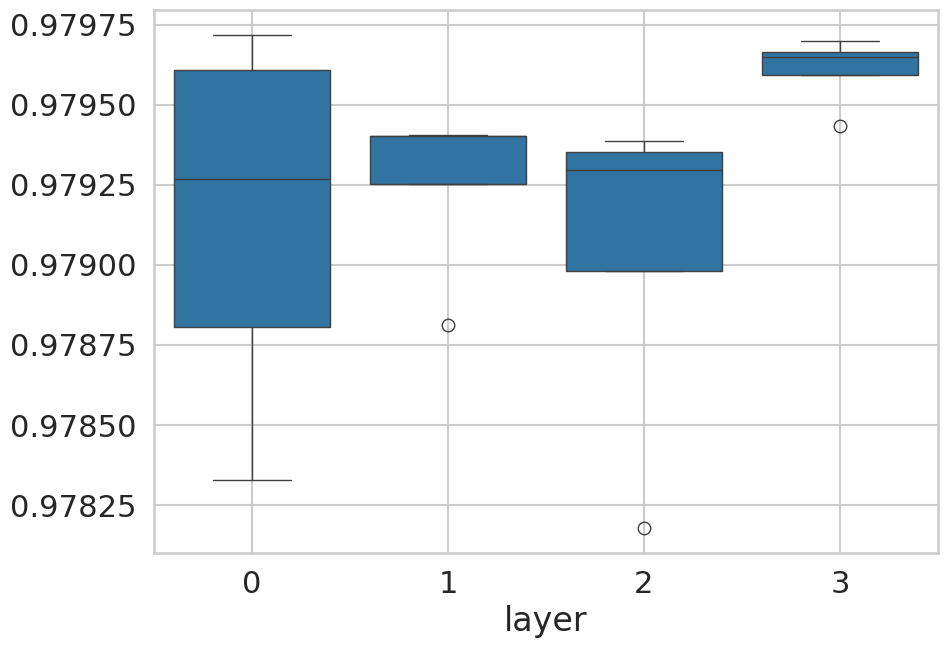} \end{minipage}
\hfill
\begin{minipage}{\imgswidth} \centering \includegraphics[width=\linewidth]{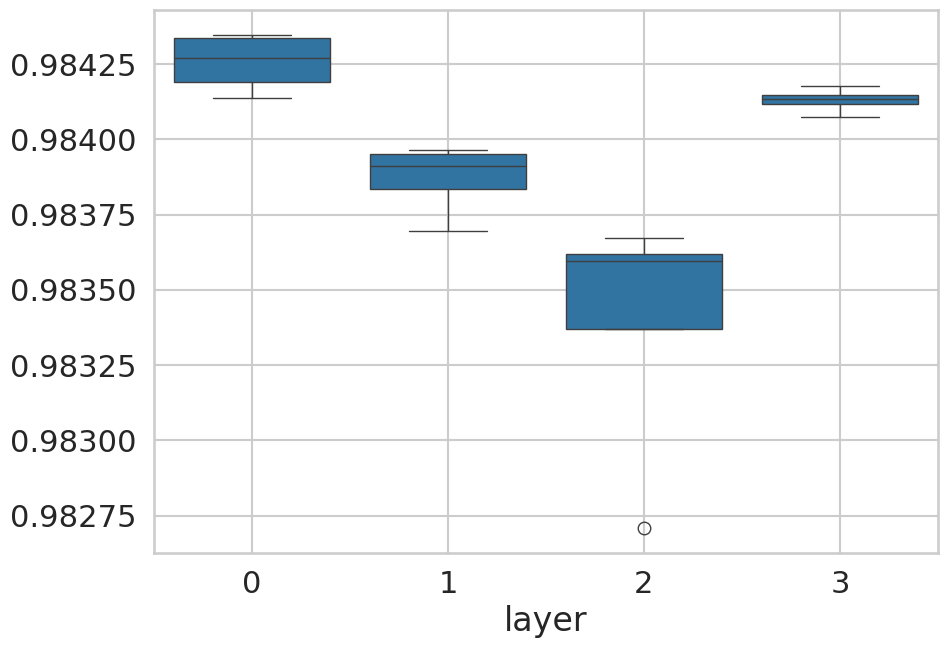} \end{minipage}
\hfill
\begin{minipage}{\imgswidth} \centering \includegraphics[width=\linewidth]{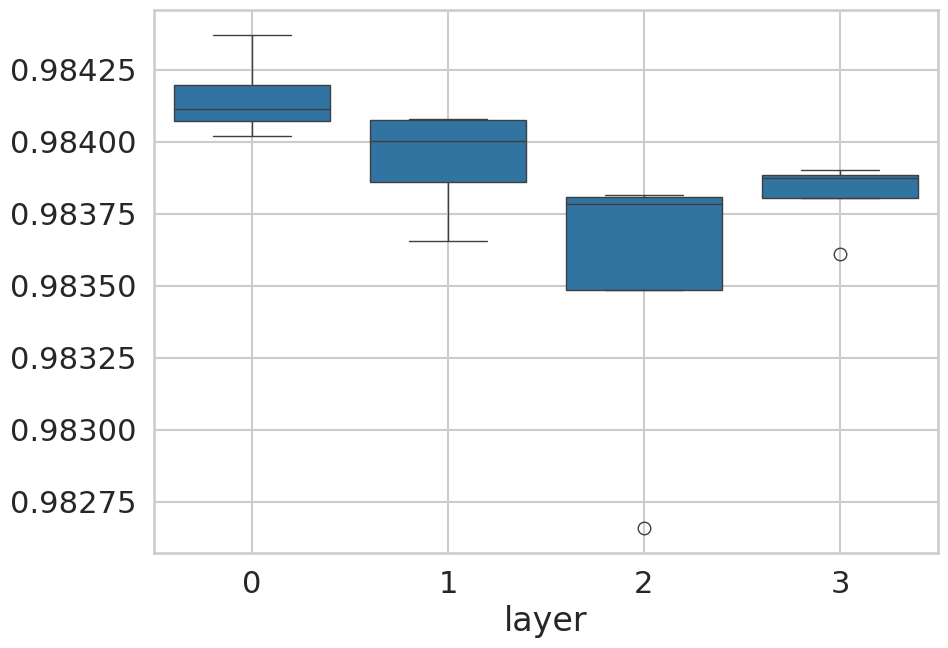} \end{minipage}
\hfill
\begin{minipage}{\imgswidth} \centering \includegraphics[width=\linewidth]{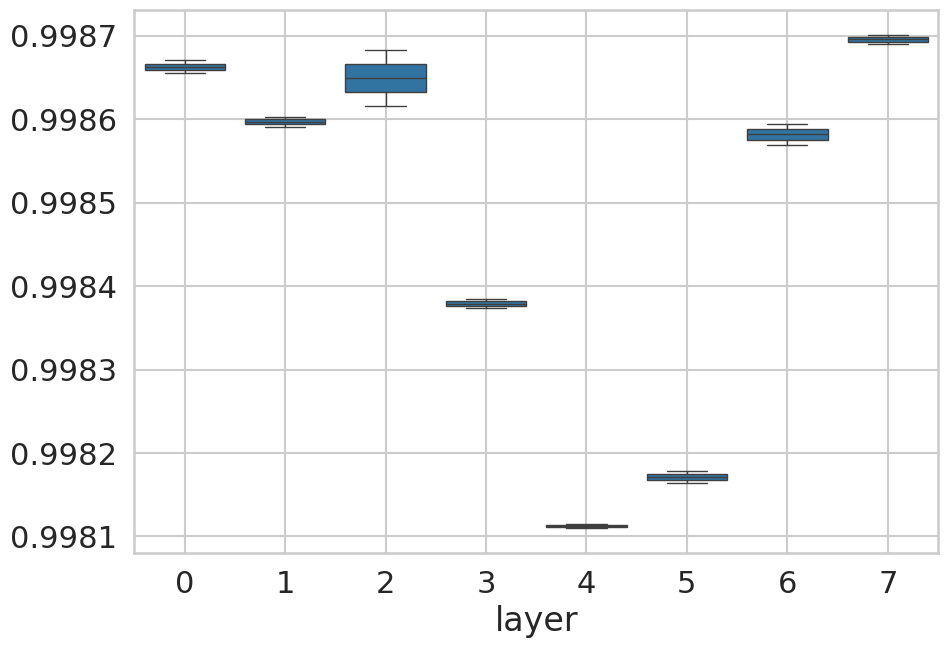} \end{minipage}
\par\vspace{2.2em}

\begin{minipage}{\labelswidth} \centering \textbf{i} \end{minipage}
\hfill
\begin{minipage}{\imgswidth} \centering \includegraphics[width=\linewidth]{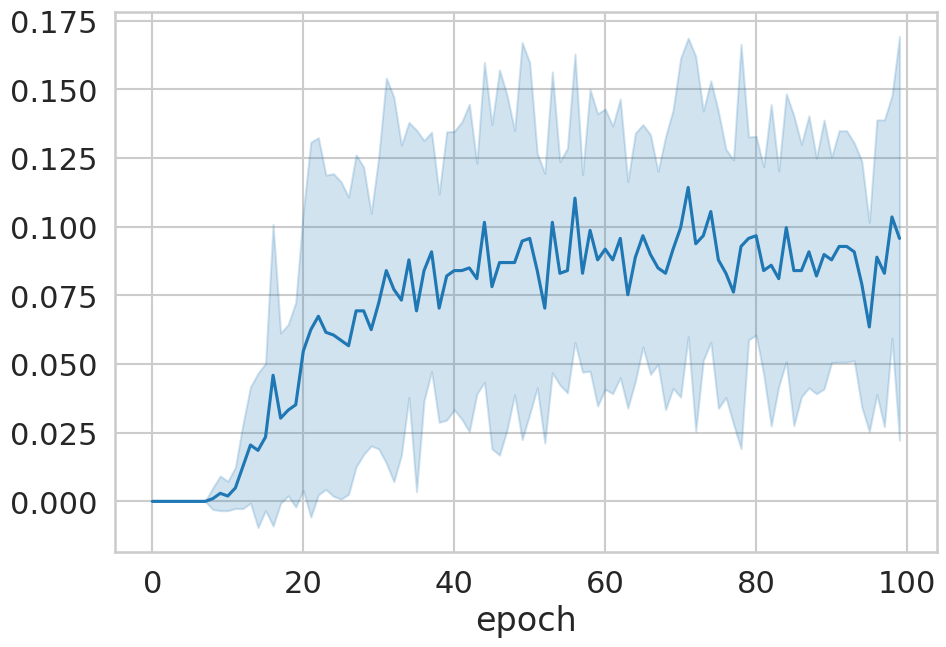} \end{minipage}
\hfill
\begin{minipage}{\imgswidth} \centering \includegraphics[width=\linewidth]{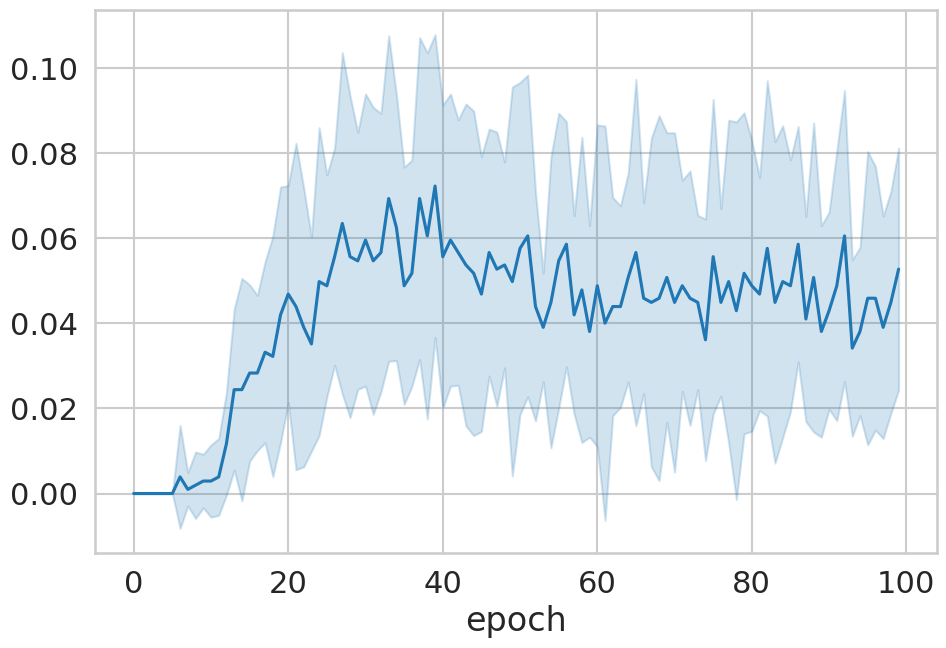} \end{minipage}
\hfill
\begin{minipage}{\imgswidth} \centering \includegraphics[width=\linewidth]{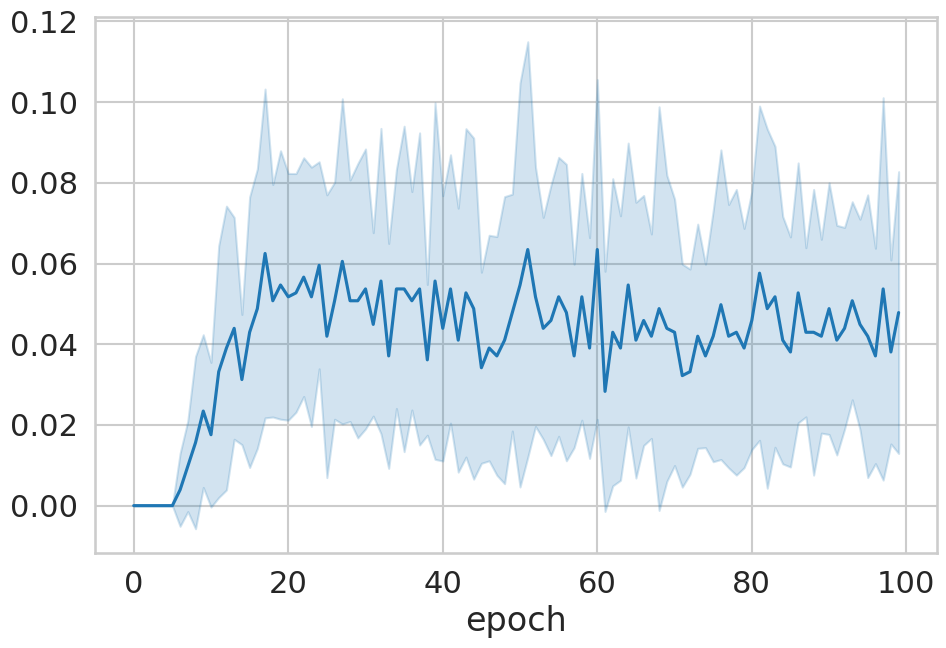} \end{minipage}
\hfill
\begin{minipage}{\imgswidth} \centering \includegraphics[width=\linewidth]{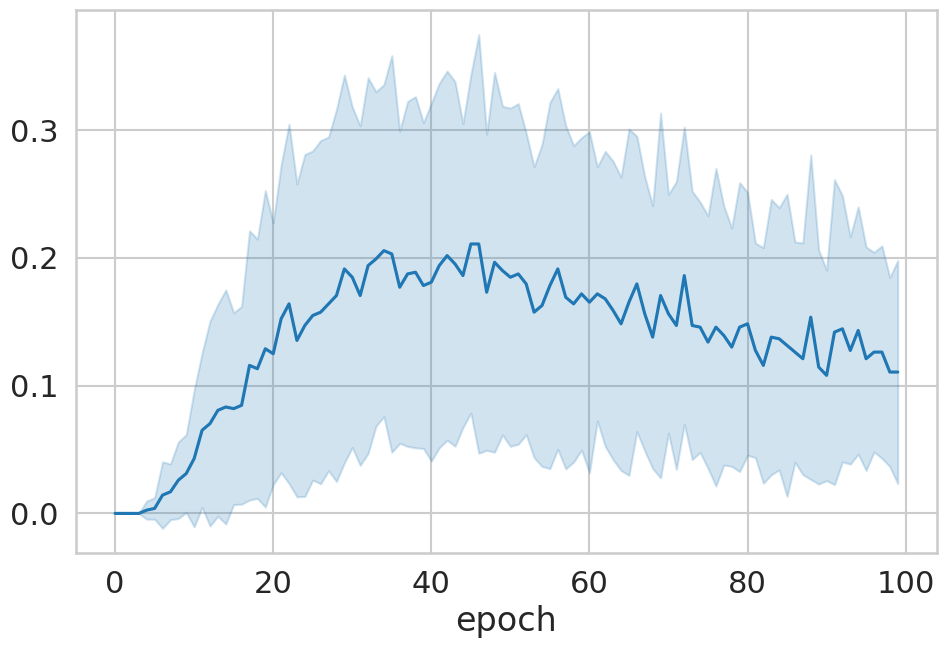} \end{minipage}
\par\vspace{2.2em}

\begin{minipage}{\labelswidth} \centering \textbf{j} \end{minipage}
\hfill
\begin{minipage}{\imgswidth} \centering \includegraphics[width=\linewidth]{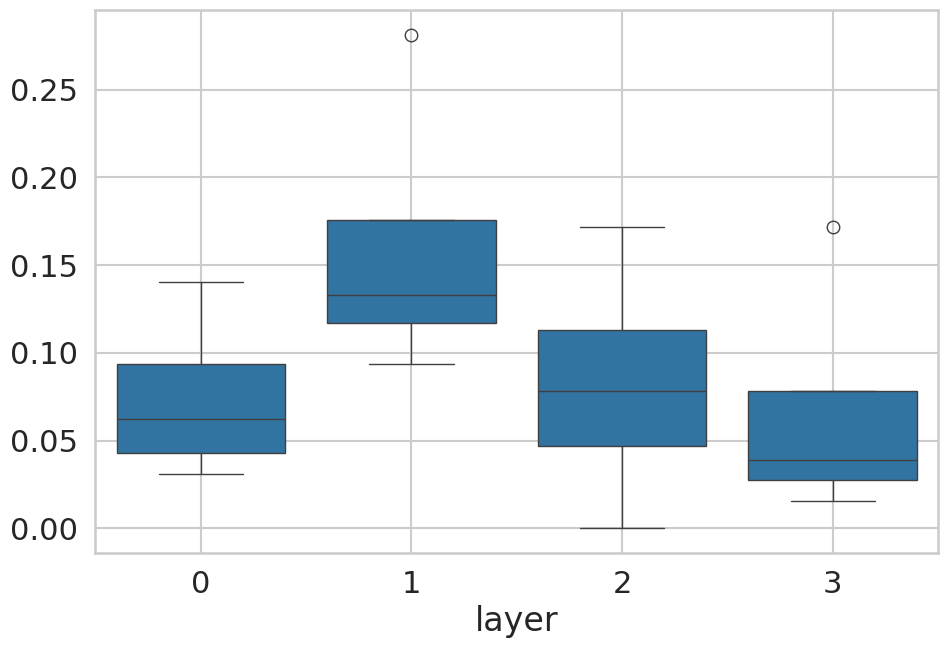} \end{minipage}
\hfill
\begin{minipage}{\imgswidth} \centering \includegraphics[width=\linewidth]{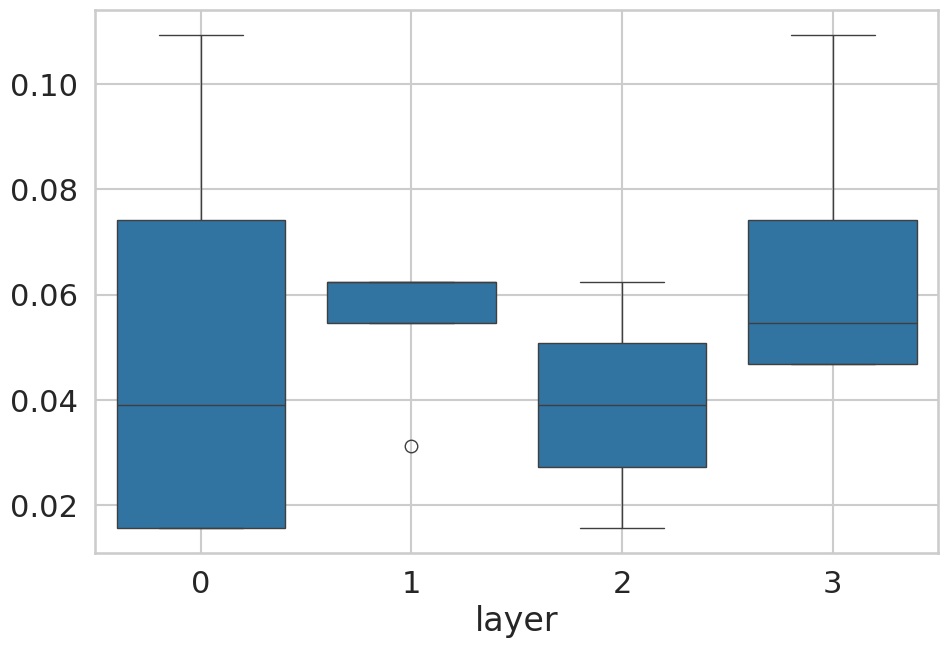} \end{minipage}
\hfill
\begin{minipage}{\imgswidth} \centering \includegraphics[width=\linewidth]{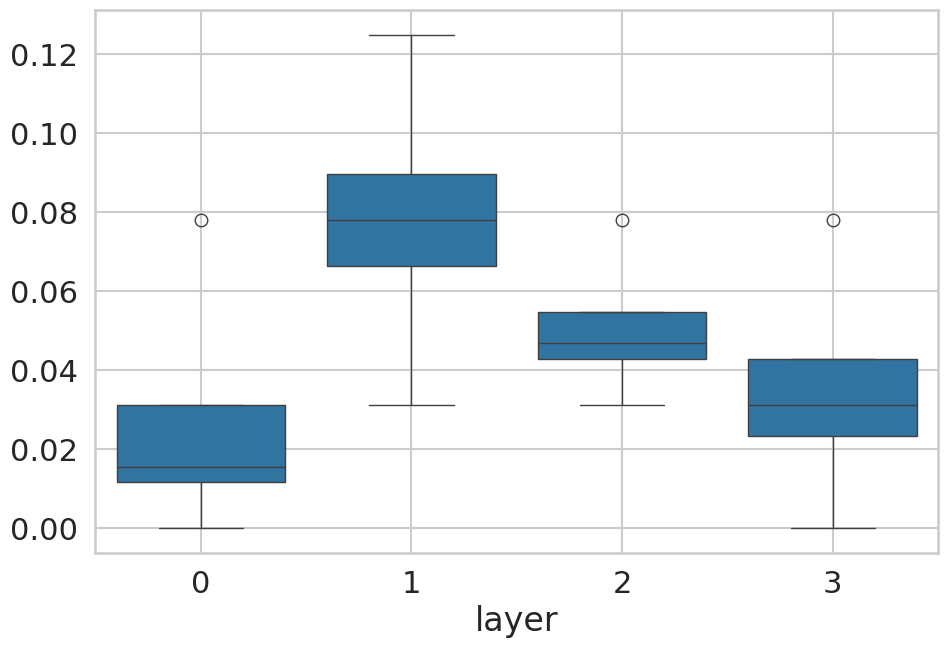} \end{minipage}
\hfill
\begin{minipage}{\imgswidth} \centering \includegraphics[width=\linewidth]{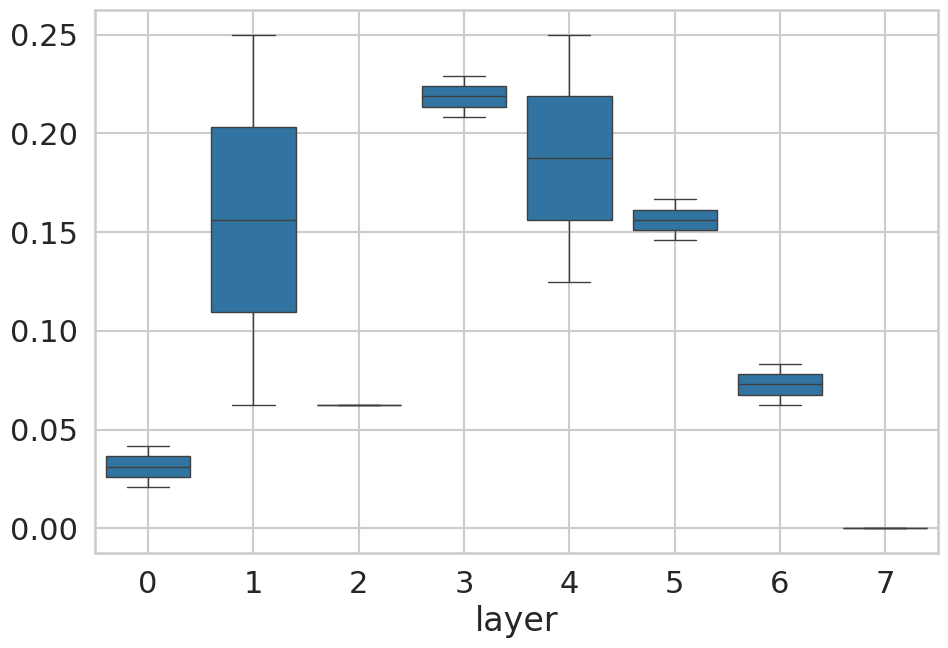} \end{minipage}
\par\vspace{2.2em}

\begin{minipage}{\labelswidth} \centering \textbf{k} \end{minipage}
\hfill
\begin{minipage}{\imgswidth} \centering \includegraphics[width=\linewidth]{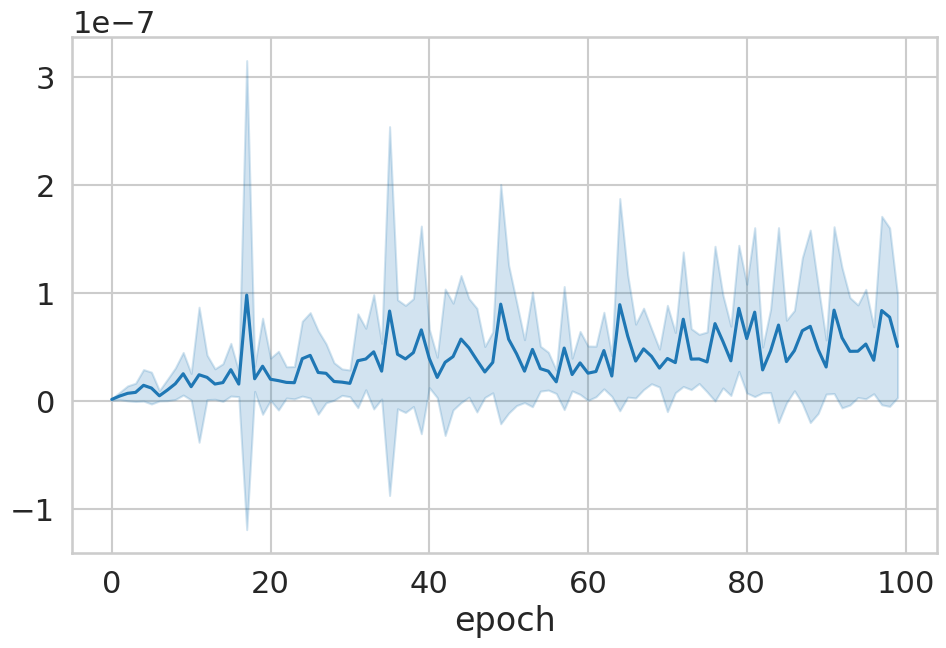} \end{minipage}
\hfill
\begin{minipage}{\imgswidth} \centering \includegraphics[width=\linewidth]{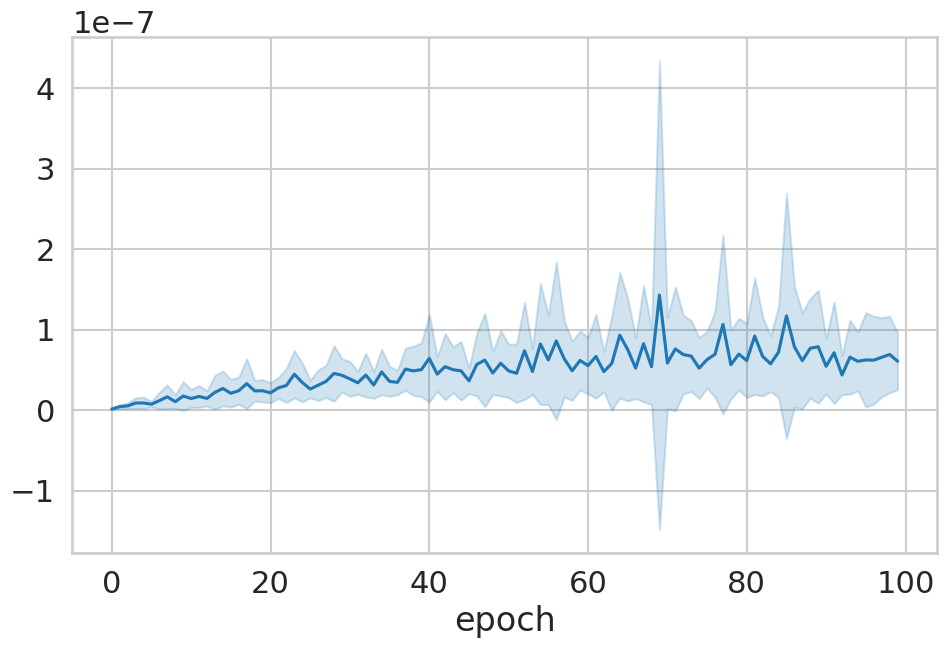} \end{minipage}
\hfill
\begin{minipage}{\imgswidth} \centering \includegraphics[width=\linewidth]{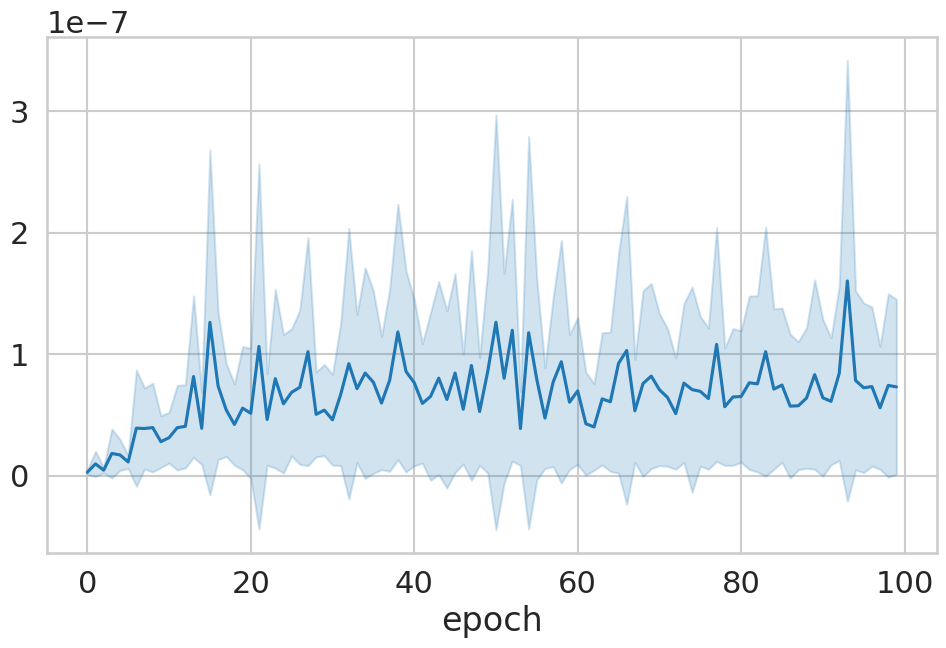} \end{minipage}
\hfill
\begin{minipage}{\imgswidth} \centering \includegraphics[width=\linewidth]{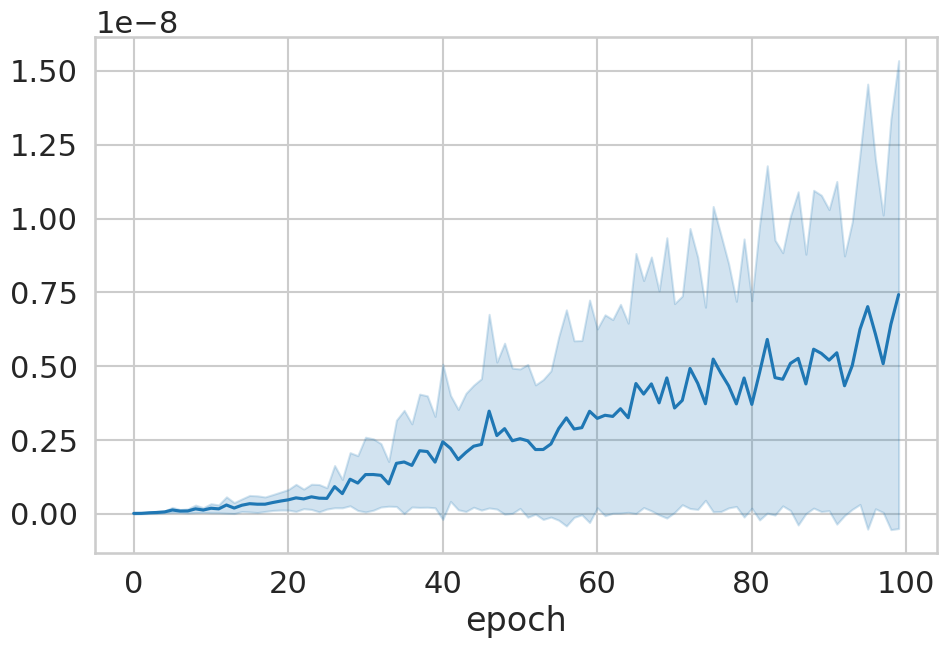} \end{minipage}
\par\vspace{2.2em}

\begin{minipage}{\labelswidth} \centering \textbf{l} \end{minipage}
\hfill
\begin{minipage}{\imgswidth} \centering \includegraphics[width=\linewidth]{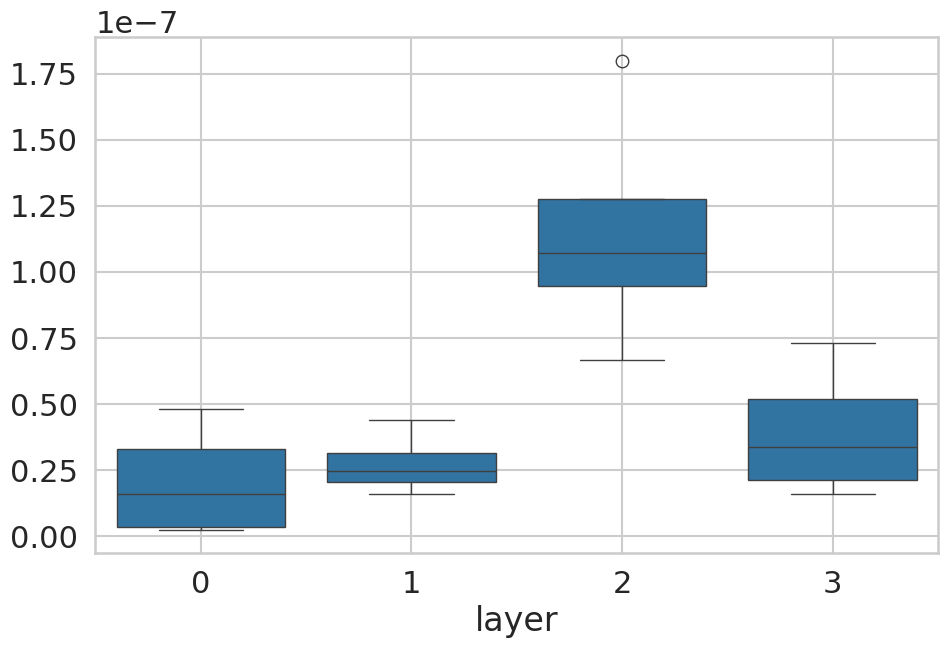} \end{minipage}
\hfill
\begin{minipage}{\imgswidth} \centering \includegraphics[width=\linewidth]{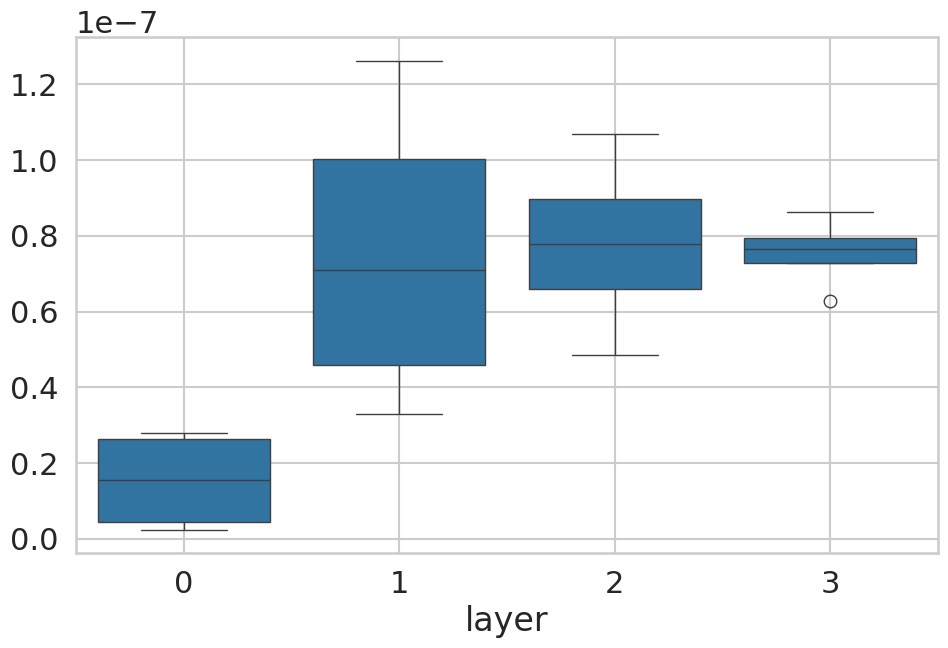} \end{minipage}
\hfill
\begin{minipage}{\imgswidth} \centering \includegraphics[width=\linewidth]{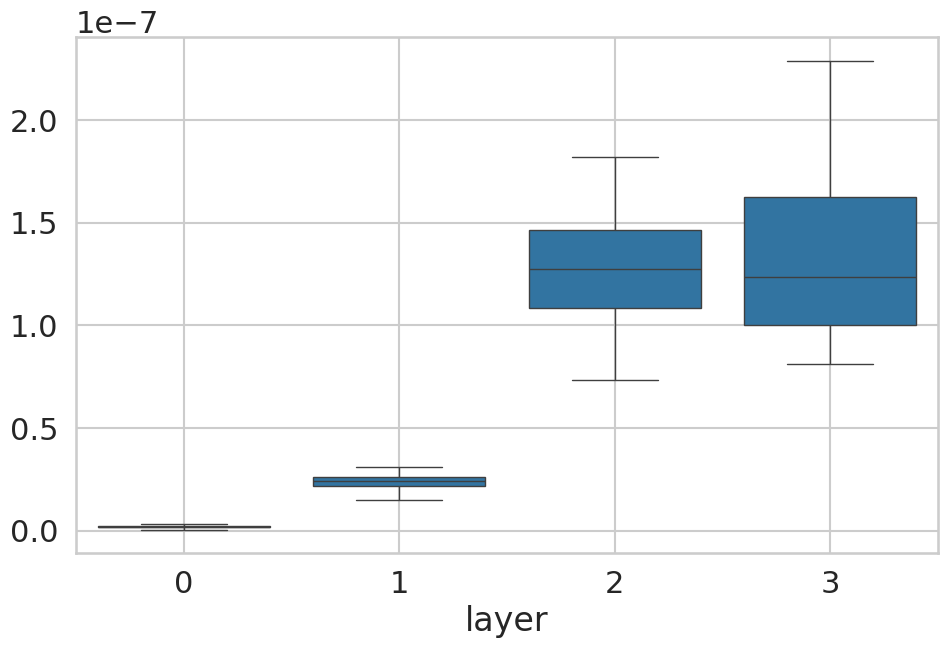} \end{minipage}
\hfill
\begin{minipage}{\imgswidth} \centering \includegraphics[width=\linewidth]{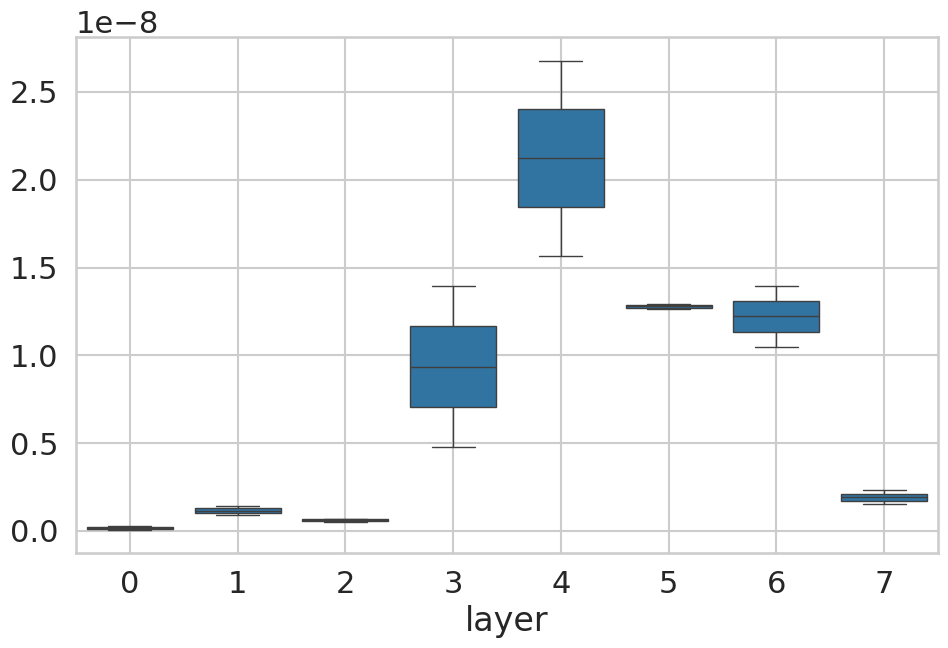} \end{minipage}
\par\vspace{0.5em}

\caption{SVDA interpretability and structural attention diagnostics across datasets and interpretability indicators (part 2 of 2): (g) Selectivity Index evolution per epoch; (h) Selectivity Index per layer; (i) Spectral Sparsity evolution per epoch; (j) Spectral Sparsity per layer; (k) Perturbation Robustness evolution per epoch; (l) Perturbation Robustness per layer.}
\label{fig:svda_grid_b}
\end{figure*}

\figurename~\ref{fig:svda_grid_a} and \figurename~\ref{fig:svda_grid_b} present the trajectories of interpretability indicators across training epochs, along with boxplot summaries per transformer layer, for all datasets over 100 training epochs.

\textbf{Temporal evolution.}
Across all datasets, spectral entropy exhibits a consistent downward trend, indicating that the attention spectra become increasingly compact over time. This reflects an emergent concentration of energy in fewer spectral components, aligned with our interpretability objectives. Effective rank similarly decreases, supporting the formation of low-rank attention maps. In contrast, angular alignment increases steadily during training, signaling stronger directional coherence between queries and keys---a marker of improved semantic alignment in the attention structure.

The selectivity index remains high (near 1) with a mild downward drift, suggesting stable but slightly more distributed attention allocation. Spectral sparsity initially rises, then stabilizes into a noisy low-level plateau after approximately 50 epochs (values between 0.04 and 0.1). This indicates that while a small fraction of singular values is suppressed, most retain significance. Perturbation robustness gradually increases, though with small absolute changes (on the order of $10^7$), indicating a mild improvement in resilience to spectral perturbations as training progresses.

\textbf{Layerwise behavior.}
Boxplot analyses across layers reveal substantial variation. For spectral entropy and effective rank, early layers tend to show broader interquartile ranges, while deeper layers converge toward tighter, more stable medians. Angular alignment and perturbation robustness exhibit pronounced median shifts across layers, suggesting hierarchical differentiation in attention structure. The selectivity index remains consistently high across layers, though its spread varies, indicating the coexistence of highly focused and more diffusely distributed heads within each layer.

Overall, this dual-axis analysis---temporal and structural---demonstrates that SVDA fosters not only interpretable attention dynamics over time but also distinctive attention behaviors across layers. These traits enhance model transparency without compromising performance.

SVDA-enabled Transformers consistently match the classification accuracy of standard attention models across all datasets, while offering markedly improved structural interpretability. The six diagnostic metrics jointly reveal that SVDA promotes a spectrum-aware and geometrically grounded attention mechanism---favoring compactness, semantic alignment, and robustness.

These benefits are especially evident in larger datasets such as ImageNet-100, where deeper layers demonstrate sharper metric differentiation and clearer functional specialization. This suggests that SVDA scales effectively and accentuates its interpretability gains in more complex learning settings.

\section{Discussion}

The empirical and structural results presented across multiple datasets establish SVDA as a viable alternative to conventional attention, balancing accuracy with interpretability. Unlike prior approaches that treat attention as an opaque black box, SVDA imposes geometric structure and spectral regularity, yielding attention maps that are more focused, semantically organized, and analyzable. This structure is not merely aesthetic but has practical implications for model robustness, generalization, and failure diagnosis.

Importantly, SVDA’s interpretability metrics are currently descriptive: they help researchers understand the model but do not yet influence training objectives. This represents both a strength---allowing transparent post hoc analysis---and a limitation, as future work must explore how to make these signals prescriptive. Additionally, while SVDA introduces minor computational overhead during training due to orthogonality constraints and spectral modulation, it does not affect inference efficiency.

These findings suggest several promising directions: integrating SVDA into more complex Transformer-based architectures (e.g., hierarchical or multi-scale models), extending the framework to other domains (e.g., segmentation, time-series, or multimodal fusion), and using SVDA metrics to drive self-regularizing attention during training. By offering a path toward explainable and structured attention, SVDA contributes to the broader agenda of making AI systems more transparent, trustworthy, and controllable.

\section{Conclusion}

We have presented Spectral-Value-Decomposed Attention (SVDA), a geometrically grounded reformulation of self-attention designed to enhance the interpretability and structure of attention mechanisms in Vision Transformers. By combining $\ell_2$-normalized query/key projections with a learned diagonal spectral modulation matrix, SVDA imposes an explicit separation between directional alignment and spectral importance, leading to more focused and semantically organized attention maps.

Empirical results across four benchmark datasets---FashionMNIST, CIFAR-10, CIFAR-100, and ImageNet-100---demonstrate that SVDA matches the classification accuracy of standard Vision Transformers while providing substantially richer interpretability. This is quantified through six diagnostic indicators: spectral entropy, effective rank, spectral sparsity, angular alignment, selectivity index, and perturbation robustness. These metrics reveal consistent improvements in attention sharpness, head specialization, and resilience to input perturbations.

SVDA offers a scalable and architecture-compatible solution for embedding interpretability into attention-based models, without compromising computational feasibility or predictive performance. Future work will investigate its integration into pretrained backbones, extension to temporal and multimodal settings, and potential for guiding prescriptive learning strategies through attention structure regularization.

\bibliographystyle{unsrt}  
\bibliography{references}  


\end{document}